%% file: main_arxiv.tex
\documentclass{article}
\usepackage{microtype}
\usepackage{graphicx}
\usepackage{subfigure}
\usepackage{booktabs}
\usepackage{hyperref}
\usepackage{amsmath}
\usepackage{amssymb}
\usepackage{amsfonts}
\usepackage{mathtools}
\usepackage{amsthm}
\usepackage{dsfont}
\usepackage{enumitem}
\usepackage[capitalize,noabbrev]{cleveref}
\usepackage[bbgreekl]{mathbbol}
\usepackage{tikz}
\usepackage{natbib}
\usepackage{fullpage}
\usepackage{algorithm}
\usepackage{algcompatible}

\theoremstyle{plain}
\newtheorem{theorem}{Theorem}[section]
\newtheorem{proposition}[theorem]{Proposition}
\newtheorem{lemma}[theorem]{Lemma}

\theoremstyle{definition}
\newtheorem{definition}[theorem]{Definition}

\theoremstyle{remark}

\DeclareMathOperator*{\argmax}{arg\,max}

\DeclareMathOperator*{\E}{\mathds{E}}

\DeclareMathOperator{\sign}{\mathrm{sign}}
\DeclareMathOperator{\corr}{\mathrm{corr}}
\DeclareMathOperator{\VCB}{VC(\mathcal{B})}
\DeclareSymbolFontAlphabet{\mathbb}{AMSb}
\DeclareSymbolFontAlphabet{\mathbbl}{bbold}

\title{Sample-Optimal Agnostic Boosting with Unlabeled Data}
\author{
    Udaya Ghai\\
    Amazon\\
    \texttt{ughai@amazon.com}
    \and
    Karan Singh\\
    Tepper School of Business\\
    Carnegie Mellon University \\
    \texttt{karansingh@cmu.edu}
}
\date{}

\begin{document}
\maketitle

\input{icml/abstract}
\input{icml/intro}
\input{icml/setting}
\input{icml/algorithm}
\input{icml/variance}
\input{icml/shift}
\input{icml/applications}

\input{icml/experiments}

\bibliography{main}
\bibliographystyle{plainnat}

\newpage
\appendix
\section*{Appendix}
\input{icml/app-var}

\input{icml/experiment_app}
\input{icml/app_applications}

\end{document}

%% file: icml/abstract.tex
\begin{abstract}   
Boosting provides a practical and provably effective framework for constructing accurate learning algorithms from inaccurate {\em rules of thumb}. It extends the promise of sample-efficient learning to settings where direct Empirical Risk Minimization (ERM) may not be implementable efficiently. In the realizable setting, boosting is known to offer this computational reprieve without compromising on sample efficiency. However, in the agnostic case, existing boosting algorithms fall short of achieving the optimal sample complexity. 

This paper highlights an unexpected and previously unexplored avenue of improvement: unlabeled samples. We design a computationally efficient agnostic boosting algorithm that matches the sample complexity of ERM, given polynomially many additional unlabeled samples. In fact, we show that the total number of samples needed, unlabeled and labeled inclusive, is never more than that for the best known agnostic boosting algorithm -- so this result is never worse -- while only a vanishing fraction of these need to be labeled for the algorithm to succeed. This is particularly fortuitous for learning-theoretic applications of agnostic boosting, which often take place in the distribution-specific setting, where unlabeled samples can be availed for free. We detail other applications of this result in reinforcement learning.
\end{abstract}

%% file: icml/intro.tex
\section{Introduction}
The methodology of boosting, starting with the work of \citet{schapire1990strength}, has deep roots both in the practice and theory of machine learning. In practice, combining classifiers trained on adaptively weighted data points highlighting past mistakes has proven to be a powerful idea. In theory, boosting provides a highly general and provably efficient framework to convert learning algorithms that are ever so slightly better than random into accurate learners. The historical origins of boosting began with the realization that although Empirical Risk Minimization (ERM), that is, the act of choosing the hypothesis that best fits the observed data, provides a general scheme for statistical learning, for many hypothesis classes it may not be approachable in terms of computational efficiency. Often instead, in theory and practice, it is possible to quickly construct weakly accurate {\em rules-of-thumb} (or weak learners). Can one use such weak learners to construct an accurate learning algorithm?\footnote{A part of the original motivation here was that such an aggregation of hypotheses would no longer be in-class (in modern parlance, {\em improper}) and hence could potentially circumvent hardness. In fact, \citet{kearns1994cryptographic} were interested in {\em representation-independent} hardness of learning, i.e., hardness that persists independently of algorithmic parametrization.} Boosting provides an affirmative answer to this question. A natural follow-up is: {\em Does the computational reprieve that boosting provides come at a cost?}

Here, we are primarily concerned with sample efficiency. In the {\em realizable} case, that is, when a perfect classifier exists, the general answer is no. This serves to reaffirm our faith in realizable boosting. The celebrated Adaboost algorithm \citep{freund1997decision}, for instance, achieves the same (near-optimal) sample complexity as ERM. In fact, \citet{green2022optimal} recently showed that for a carefully designed variant the equivalence holds even up to logarithmic terms. 

Let us now move to the {\em agnostic} case, where we make no assumptions about the conditional distribution of labels given features; in many ways, this has become the default stance in machine learning. Firstly, the classical notion of realizable weak learning requires that weak learners attain a $0/1$ loss strictly better than half, or equivalently, a correlation strictly better than zero on signed labels. However, such notions are unattainable in the agnostic setting. Early works on agnostic boosting using weak learning definitions that mended this issue did not produce results that were additively competitive with the underlying hypothesis class, thus a direct comparison to ERM remained elusive. Following this, \citet{kanade2009potential} gave a computationally efficient agnostic boosting that in fact achieves $\varepsilon$-excess population loss in comparison to the best-in-class hypothesis, mirroring the guarantee ERM provides when coupled with generalization bounds (e.g., in \citet{shalev2014understanding}). However, the sample complexity of their boosting algorithm scales as ${(\log |\mathcal{H}|)}/{\varepsilon^4}$, where $\mathcal{H}$ is the (let us say for now, finite\footnote{Purely for ease of presentation, in the introduction, we stick to finite classes and drop dependencies in the weak learning edge and the failure probability. In fact, our results hold for infinite VC classes. See \Cref{thm:main}}) hypothesis class. A recent result of \citet{ghai2024sampleefficient} provides a booster that requires ${(\log |\mathcal{H}|)}/{\varepsilon^3}$ samples. {\em However, both exhibit significantly higher sample complexity than the familiar $(\log |\mathcal{H}|)/\varepsilon^2$ required for ERM, which is also optimal.}

Although agnostic boosting has seen considerable recent progress both in methodology, e.g., through extensions to online \citep{brukhim2020online} and multiclass \citep{raman2022online} settings, and in finding new applications \citep{brukhim2022boosting, kothari2018improper}, this limitation persists.

In this work, we highlight an unexpected and unexplored avenue of improvement: {\em unlabeled} samples. As usual, to measure the statistical efficiency of the learner, we shall keep track of the number of labeled samples it needs. However, we imbue the learner with the ability to additionally draw polynomially many unlabeled samples from the feature distribution. Under this stipulation, our main result is a computationally efficient agnostic boosting algorithm with (labeled) sample complexity scaling as $(\log |\mathcal{H}|)/\varepsilon^2$, and therefore for the first time matching that of ERM. 

This is fortuitous and relevant both on practical and theoretical fronts. In practice, it is widely acknowledged that, in many domains, unlabeled samples are substantially cheaper to gather than labeled ones. Additionally, many applications of agnostic boosting take place in settings where unlabeled samples can be drawn for free; for these, even closing the aforementioned gap will obtain results that are no better than what we get. We offer further refinements and applications of our main result, as we detail next.

\subsection{Overview of contributions and techniques}

{\bf Sample-optimal boosting with unlabeled data.} Our first and main result is a computationally efficient agnostic boosting algorithm that produces a classifier with $\varepsilon$ excess error, given $(\log |\mathcal{H}|)/ \varepsilon^2$ labeled and $(\log |\mathcal{H}|)/ \varepsilon^4$ unlabeled samples. Our labeled sample complexity is essentially optimal.

Using unlabeled data to accelerate boosting itself is a novel idea. Our work also differs qualitatively, in that it does not require additional assumptions, compared to semi-supervised learning methods \citep{chapelle2006semi}, which often require a tight clustering of data or smoothness of labels.

Previous works \citep{kanade2009potential, ghai2024sampleefficient} on agnostic boosting considered the Madaboost potential \citep{domingo2000madaboost} or a closely related variant:
$$ \phi_\textrm{MADA}(z, y) = \begin{cases}
	e^{-zy} & \text{ if } zy \geq 0, \\
	1-zy &\text{ if } zy < 0.
\end{cases}$$
Our main innovation is to design a new {\em bivariate} potential (\Cref{eq:new_potential}) whose derivative is decomposable as one term exclusively dependent on the label, and the other exclusive involving the features. Thus, these parts can be estimated separately with labeled and unlabeled data.

This non-conforming potential also brings algorithmic changes. For example, previous works restrict themselves to {\em conservative} potentials, those that do not under-weight mislabeled points. In fact, the previous labeled-unlabeled separation property is incompatible with this requirement. Also, unlike previous work, our algorithm reuses the entire set of labeled examples across all rounds of boosting without paying for uniform convergence on the {\em boosted} class.

{\bf Improved unlabeled sample efficiency.} Applying techniques from \citet{ghai2024sampleefficient} selectively to unlabeled data, we further reduce the number of unlabeled samples needed to $\log (|\mathcal{H}|)/\varepsilon^3$. With this result, our total number of samples (labeled and unlabeled) matches the number of labeled samples needed by the previous best result, yet for us, only a vanishing fraction of these needs to be labeled.

{\bf Resilience to covariate shift.} Unlike realizable learning where every sample helps to narrow down to the correct classifier, agnostic learning is often brittle against changes in the covariate distribution, since no classifier is the best in all regions of the feature space. However, we show that if there is a mismatch between the labeled and unlabeled distributions available to learner, our learner still succeeds in learning an {\em arbitrarily accurate} classifier as long as these distributions have the same support on the feature space. Moreover, the labeled sample complexity is unaffected.

{\bf Applications.} We apply our results to boosting for reinforcement learning, where we reduce the required number of reward-annotated episodes, and learning halfspaces.

\subsection{Related work}
The early theory of boosting was developed in a sequence of papers \citep{freund1995boosting, schapire1998improved, bartlett1998boosting} starting with \citet{schapire1990strength} leading to a breakthrough in the form of Adaboost \citep{freund1997decision}. Even earlier the possibility of boosting was posed in \citet{kearns1988learning, kearns1994cryptographic}. A comprehensive survey can be found in \citet{schapire2013boosting}. Early boosting algoritms were quite sensitive to noise \citep{long2008random,domingo2000madaboost}. To mitigate this, boosting was then studied in the random classification noise model \citep{diakonikolas2021boosting, kalai2003boosting}. Agnostic boosting started with \citet{ben2001agnostic, mansour2002boosting, kalai2008agnostic, kale2007boosting, chen2016communication} where new types of weak learners suitable for the agnostic setting were defined. \citet{kanade2009potential} (see also \citet{feldman2009distribution}) gave weak learning definition that led to additive excess error guarantees. Boosting in online setting is also well studied \citep{beygelzimer2015optimal, chen2012online, jung2017online,brukhim2020online, raman2022online, hazan2021boosting}. See \citet{alon2021boosting,green2022optimal,lyu2024cost} for more recent work.

%% file: icml/setting.tex
\section{Problem setting}
We consider the fundamental setting of binary classification where $\mathcal{X}$ represents the set of feature descriptions, and the possible labels are $\{\pm 1\}$. There is an underlying joint distribution $\mathcal{D}$ over $\mathcal{X}\times \{\pm 1\}$, which, while unknown to the learner, is crucial to determining its performance. Let $\mathcal{D}_\mathcal{X}$ be the marginal distribution $\mathcal{D}$ induces on the feature space $\mathcal{X}$. Given any binary classifier $h:\mathcal{X} \to \{\pm 1\}$, its success on $\mathcal{D}$ can be measured as
$$\corr_\mathcal{D}(h) = \E_{(x,y)\sim \mathcal{D}} \left[yh(x)\right].$$
Correlation can readily be translated to the more commonly used metric, the $0/1$-loss $l^{0/1}_{\mathcal{D}}(h) = \E_{(x,y)\sim \mathcal{D}} \left[\mathbf{1}_{y\neq h(x)}\right]$, as $\corr_\mathcal{D}(h) = 1-2l^{0/1}_{\mathcal{D}}(h)$. 

Consider a hypothesis class $\mathcal{H}\subseteq \mathcal{X} \to \{\pm 1\}$ against which the learner aims to be competitive. The objective of the learner is to produce a hypothesis $\overline{h}:\mathcal{X}\to \{\pm 1\}$ such that with probability at least $1-\delta$,  $\corr_\mathcal{D}(\overline{h}) \geq \max_{h\in \mathcal{H}} \corr_\mathcal{D}(h) - \varepsilon$, where $\varepsilon,\delta$ are pre-specified error tolerances. A crucial remark here is that this final hypothesis may not belong to the hypotheses class $\mathcal{H}$. The learners who are given such flexibility are called {\em improper}, and all known boosting algorithms fall into this class.

\begin{definition}[Agnostic Weak Learner]\label{def:wl}
For any $\varepsilon_0, \delta_0>0$, a $\gamma$-agnostic weak learner for a hypothesis class $\mathcal{H}$ and a base class $\mathcal{B}$ draws $m(\varepsilon_0, \delta_0)$ independently and identically distributed samples from any distribution $\mathcal{D}'$ supported on $\mathcal{X}\times \{\pm 1\}$ and outputs a base hypothesis $\mathcal{W}\in \mathcal{B}$ such that with probability at least $1-\delta_0$,
$$ \corr_\mathcal{D'}(\mathcal{W}) \geq \gamma \max_{h\in \mathcal{H}}\corr_\mathcal{D'}(h) -\varepsilon_0.$$
\end{definition} 

This definition of agnostic weak learning was introduced in \citet{kanade2009potential}, where it is noted that typically $m(\varepsilon,\delta) = O({\log (|\mathcal{B}|/\delta})/{\varepsilon^2})$. Mirroring the presentation of results in \citet{kanade2009potential, brukhim2020online,  ghai2024sampleefficient}, we present the formal statement of results for fixed $\varepsilon_0, \delta_0$. In this way, the magnitudes of contribution to the final error for the boosting algorithm and the weak learner are made distinct.

Finally, a further refinement of the agnostic boosting framework is the {\em distribution-specific} setting \citep{kanade2009potential, feldman2009distribution}, in which the set of input distributions to the weak learner are constrained so that their marginals on the feature space match that of the true underlying distribution. Thus, any regularity present in the feature distribution $\mathcal{D}_\mathcal{X}$, e.g., if it follows a uniform or a Gaussian distribution, is also made available to the weak learner, which makes the design of such weak learners easier. Our results will also apply under this restriction. Like in previous work, our main algorithm works by {\em relabeling examples}; there is no need to adaptively reweigh them. This ensures that, for any sample fed to the weak learner, the overall marginal distribution follows $\mathcal{D}_\mathcal{X}$.

%% file: icml/algorithm.tex
\begin{figure}
	\centering
    \begin{tikzpicture}[scale=0.7]
        \draw[color=gray, thin, dotted] (-2.1, -0.6) grid (4.1, 3.1);
    
        \draw[->] (-2.2,0) -- (4.3,0);
        \draw[->] (0,-0.8) -- (0,3.2);
    
        \draw[color=red, domain=-2:2, smooth] plot (\x, {(\x < -1)*(-\x-0.5) + (\x >= -1)*(\x <= 1)*(0.5*\x*\x) + (\x > 1)*(\x-0.5) }) node[right] {$\psi(z)$};
        \draw[color=blue, domain=-2:2, smooth] plot (\x, {0.5*((\x < -1)*(-\x-0.5) + (\x >= -1)*(\x <= 1)*(0.5*\x*\x) + (\x > 1)*(\x-0.5) - \x)}) node[yshift=-0.2em, right] {$\phi(z, 1)/2$};
        \draw[color=olive, domain=-2:2, smooth] plot (\x, {(\x < 0)*(1 - \x) + (\x >= 0)*(e^-\x)}) node[yshift=0.4em, right] {$\phi_{\textrm{MADA}}(z,1 )$};
    \end{tikzpicture}
	\caption{The potential $\color{blue}\phi(z,1)/2$  compared against the Madaboost potential $\color{olive} \phi_\textrm{MADA}(z,1)$ \cite{domingo2000madaboost}, along with the the Huber loss $\color{red} \psi(z)$.}
    \label{fig:potential}
\end{figure}

\section{The algorithm and the main result}\label{sec:main}
Our main result and its proof are entirely contained in this section. We describe some notation and essential algorithmic ingredients below. 

\paragraph{Notation.}
Let $h^* = \argmax_{h\in \mathcal{H}} \corr_\mathcal{D}(h)$ be the best-in-class hypothesis. For brevity of notation, for any finite set $D\subseteq \mathcal{X}\times\{\pm 1\}$, we denote the empirical average over it by $\widehat{\E}_{D} [\cdot] =  \frac{1}{|D|}\sum_{(x,y)\in D} (\cdot)$. We define $\sign(z)$ as $1$ if $z\geq 0$ and $-1$ otherwise. For a real-valued function $f$, we take $\sign f$ to mean its precomposition with $\sign$.

\paragraph{Potential function.} Consider the potential function 
\begin{equation}
    \phi(z,y) = \psi(z) - yz~, \label{eq:new_potential}
\end{equation} where $\psi(z)$ is the Huber loss \citep{huber1992robust}: $$\psi(z) = \begin{cases}
    |z| - \frac{1}{2} & \text{if } |z| > 1, \\
    \frac{1}{2}z^2 & \text{if } |z| \leq 1.
\end{cases}$$ 

Since we never differentiate $\phi(z,y)$ with respect to $y$, let $\phi'(z,y)=\frac{\partial\phi(z,y)}{\partial z}$ and $\phi''(z,y)=\frac{\partial^2\phi(z,y)}{\partial^2 z}$.
To measure the progress of any {\em real-valued} $H:\mathcal{X}\to \mathbb{R}$, consider the population potential \[\Phi_\mathcal{D}(H) = \E_{(x,y)\sim \mathcal{D}} \left[\phi(H(x), y)\right].\]
\[\Phi'_\mathcal{D}(H,h) = \E_{(x,y)\sim \mathcal{D}} \left[\phi'(H(x), y) h(x)\right]\]
The quantity $\Phi'_\mathcal{D}(H,h)$ is the directional derivative of $\Phi_\mathcal{D}(H)$ on $h$.

\paragraph{Description of the Algorithm.}
\Cref{alg:rev} roughly follows the potential based boosting framework of \citet{kanade2009potential}. The algorithm simulates the process where the true label is kept with probability $-\phi'(H_t(x), y)/2$ and chosen randomly otherwise. As can be seen in \cref{fig:potential}, the probability of flipping increases monotonically in the magnitude of $yH_t(x)$, so the more certain $H_t$ is of the correct label, the closer to random the label will be in $\mathcal{D}_t$ for the weak learner. Since predicting on random labels is impossible, this randomized relabeling increases the relative importance of data that $H_t$ does not predict accurately.

The main trick we employ in this work is that the potential $\phi(z,y)$ in \eqref{eq:new_potential} is split into two parts such that the first part $\psi(z)$ has \emph{no dependence on the label}, and hence can be estimated via unlabeled examples. The second part $-yz$ is \emph{linear} in $z$, hence the derivative has no dependence on $z$. As a result estimating this does not depend on $H_t$, but just a weak hypothesis. Since this is a simple class, the samples required for this estimation are small and we can use uniform convergence to assure concentration across all boosting rounds. The formal concentration argument can be seen in \cref{lem:concmain}. One interesting observation, is that in this construction, samples from the labeled part of the distribution are never relabeled. Line $6$ of \cref{alg:rev} can be interpreted as providing a regularization, wherein predictions on the unlabeled data are pushed towards $0$ because $p_t(x) \leq \frac{1}{2}$ if and only if $H_t(x) \geq 0$.

The relabeling is designed so correlation on the relabeled distribution corresponds to the derivative of a population potential (see \cref{lem:concmain}). As such, a weak learner produces a hypothesis that has nonnegligeable correlation with the (negative) functional gradient of the population potential $\Phi_{\mathcal{D}}(H_t)$. With properly chosen $\eta$, this assures a descent in potential as long as the weak learner has sufficient edge on the current distribution (Case A of \cref{thm:main} and line $10$ of \cref{alg:rev}). However, this need not be the case because we have access to an \emph{agnostic} weak learner, and it's possible that no $h \in \mathcal{B}$ performs well on $\mathcal{D}_t$. If this is the case and if $\Phi'_{\mathcal{D}}(H_t, \sign(H_t)) \geq \varepsilon$ we are in Case B of \cref{thm:main}. In this case, this condition on the derivative assures us that $h_t = - \sign H_t$ is also a descent direction (line $12$ of \cref{alg:rev}). Now, because the potential is bounded from below, only a certain number of such descent steps can occur. Eventually, we must reach a point where neither Case A nor Case B holds. Here $\Phi'_{\mathcal{D}}(H_t, \sign(H_t))$ is small and there is no $h \in \mathcal{B}$ that performs well on $\mathcal{D}_t$. Stitching these together with \cref{lem:consmain} relating the correlations to the potential gradient can be used to provide the result.
\begin{algorithm}[ht]
\begin{algorithmic}[1]
    \STATE \textbf{Inputs:} Samplers for labeled data from $\mathcal{D}$ and unlabeled data from $\mathcal{D}_\mathcal{X}$, $\gamma$-agnostic weak learning oracle $\mathcal{W}$, parameters $\eta, T, \tau, S, U, S_0, m$.
    \STATE Initialize a zero hypothesis $H_1=\mathbf{0}$.
    \STATE Sample $S$ {\em labeled} examples to create dataset $\widehat{D}$.
    \FOR{$t=1$ to $T$}
        \STATE Sample $U$ {\em unlabeled} examples to create  dataset $\widehat{D}_t$.
        \STATE Construct a resampling distribution $\mathcal{D}_t$ that chooses between steps A and B with equal probability.
            \begin{enumerate}[label=\Alph*.,itemsep=-0.2em, topsep=0.2em]
            	\item Return $(x,y)$ picked uniformly from $\widehat{D}$.
            	\item Pick $x$ uniformly from $\widehat{D}_t$, and return $(x, \widehat{y})$, where the pseudo-label $\widehat{y}$ is chosen as 
            \end{enumerate}
            $$\widehat{y} =\begin{cases}
            	+1 & \text{with probability } p_t(x) = \frac{1-\psi'(H_t(x))}{2},\\
            	-1 & \text{with remaining probability.}
            \end{cases}$$
        \STATE Sample $m$ times from $\mathcal{D}_t$ to create dataset $\widehat{D}'_t$.
        \STATE Call the weak learner on $\widehat{D}'_t$ to get $W_t=\mathcal{W}(\widehat{D}'_t)$.
        \IF{$\corr_{\widehat{D}'_t}(W_t) = \sum_{(x,\widehat{y})  \in \widehat{D}'_t} \widehat{y} W_t(x) >\tau$}
        	\STATE Update $H_{t+1} = H_t + \eta W_t/\gamma$.
        \ELSE
        	\STATE Update $H_{t+1} = H_t - \eta \sign(H_t)_t$.
        \ENDIF
    \ENDFOR
    \STATE Sample $S_0$ {\em labeled} examples to create dataset $\widehat{D}_0$.
    \STATE \textbf{Output} $\overline{h} = \argmax\limits_{h\in \{\sign(H_t):t\in [T]\}} \sum\limits_{(x,y)\in \widehat{D}_0} yh(x)$.
    \caption{Agnostic Boosting with Unlabeled Data} \label{alg:rev}
\end{algorithmic}
\end{algorithm}

\subsection{Main result on sample-optimal agnostic boosting}
\begin{theorem}[Main theorem]\label{thm:main}
For any $\varepsilon,\delta>0$, there is an instantiation of parameters such that $\eta=\mathcal{O}(\gamma^2\varepsilon)$, $T=\mathcal{O}(/\gamma^2\varepsilon^2)$, $\tau=\mathcal{O}(\gamma\varepsilon)$, $S=\mathcal{O}(\VCB/\gamma^2\varepsilon^2)$, $U=\mathcal{O}(\VCB/\gamma^2\varepsilon^2)$, $S_0=\mathcal{O}(1/\varepsilon^2)$, $m=m(\varepsilon_0,\delta_0)+\mathcal{O}(1/\gamma^2\varepsilon^2)$ for which \Cref{alg:rev} guarantees with probability $1-\delta - T\delta_0$ that \[\corr_\mathcal{D}(\overline{h}) \geq \max_{h\in \mathcal{H}} \corr_\mathcal{D}(h) - \frac{2\varepsilon_0}{\gamma} - \varepsilon.\]
During its execution, \Cref{alg:rev} makes $T=2/\gamma^2\varepsilon^2$ calls to the weak learner, and samples $S+S_0=\mathcal{O}(\VCB/\gamma^2\varepsilon^2)$ labeled samples and $TU=\mathcal{O}(\VCB/\gamma^4\varepsilon^4)$ unlabeled samples. 
\end{theorem}

\subsection{Proof of the main result}
To maintain the continuity of presentation, our organization and notation closely mirror \citet{kanade2009potential}. 
We will use the following properties of $\phi$ and $\psi$.
\begin{proposition}\label{lem:phimain}
	$|\psi'(z)|\leq 1$ and $z\psi'(z)\geq 0$ for all $z\in \mathbb{R}$. For all $y\in \{\pm 1\}$, $\phi(\cdot, y)$ is continuously differentiable, $1$-smooth, and satisfies $\phi(0, y) - \min_z \phi(z,y) \leq 1/2$. 
\end{proposition}
\begin{proof}[Proof of \Cref{lem:phimain}] 
Let us begin by noting that $\psi'(z) = \sign(z)\min\{1, |z|\}$ from which the first two properties can be seen. From this, all but the last properties follow. The last part can be verified by elementary calculations.
\end{proof}

 We show that $\Phi'_\mathcal{D}(H,h)$ can be estimated efficiently on the base class $\mathcal{B}$.

\begin{lemma}\label{lem:concmain}
	There exists a universal constant $C>0$ such that with probability $1-\delta$, for all $t\in [T], h\in \mathcal{B}\cup \{h^*\}$,
	$$\left\lvert\frac{1}{2}\Phi'_\mathcal{D}(H_t, h) + \corr_{\mathcal{D}_t}(h)\right\rvert \leq \varepsilon_{\textrm{Gen}}\coloneqq C \sqrt{\frac{\VCB + \log \frac{1}{\delta}}{\min \{S, U\}}}.$$
\end{lemma}
\begin{proof}[Proof of \Cref{lem:concmain}]
By the definition of $\mathcal{D}_t$, we have that 
\begin{align*}
\corr_{\mathcal{D}_t}(h) = \frac{1}{2}\widehat{\E}_{\widehat{D}} [yh(x)] - \frac{1}{2}\widehat{\E}_{\widehat{D}_t} [\psi'(H_t(x))h(x)],
\end{align*}
where we use the fact that Line 6.B in \Cref{alg:rev} ensures $\E[\widehat{y}|x] = -\psi'(H_t(x))$. Since $\widehat{D}$ and $\widehat{D}_t$ are composed of IID draws from $\mathcal{D}$ and $\mathcal{D}_\mathcal{X}$ respectively, we use the following uniform convergence result (e.g., see \citet{anthony2009neural}), originally due to \citet{talagrand1994sharper}.
\begin{theorem}[\cite{talagrand1994sharper}]
	Fix a hypothesis class $\mathcal{B}\subseteq \mathcal{X}\to\{\pm 1\}$, and distribution $\mathcal{D}$ over $\mathcal{X}\times \{\pm 1\}$. There is a universal constant $C\geq 0$ such that with probability $1-\delta$, for all $h\in \mathcal{B}$, it holds
		$$\left\lvert \E_{(x,y)\sim \mathcal{D}}[yh(x)] -\frac{1}{m}\sum_{i=1}^m y_i h(x_i)\right\rvert \leq C \sqrt{\frac{\VCB + \log \frac{1}{\delta}}{m}},$$
		where $\{(x_i, y_i)\}_{i\in [m]}$ are sampled IID from $\mathcal{D}$.
\end{theorem}
Hence, for some constant $C\geq 0$ we have with probability $1-\delta$ that $|\widehat{\E}_{\widehat{D}} [yh(x)] - \corr_\mathcal{D}(h)|$ and $|\widehat{\E}_{\widehat{D}_t} [\psi'(H_t(x))h(x)] - \E_{x\sim\mathcal{D}_\mathcal{X}}[\psi'(H_t(x))h(x)] |$ are at most $\varepsilon_{\textrm{Gen}}$. Since $\phi'(z,y) = \phi'(z)-y$, and hence
\[ \Phi'_\mathcal{D}(H_t,h) = \E_{x\sim \mathcal{D}_\mathcal{X}} [\psi'(H_t(x))h(x)] - \E_{(x,y)\sim \mathcal{D}} [yh(x)], \]
completes the proof. Including $h^*$ changes the VC dimension by a constant \citep{eisenstat2007vc}.
\end{proof}

Finally, we will use the following lemma that upper bounds the correlation gap, which is our ultimate concern, by the difference in the directional derivative of the potential.
\begin{lemma}\label{lem:consmain}
    For any real-valued classifier $H:\mathcal{X}\to \mathbb{R}$, we have 
    $\corr_\mathcal{D}(h^*) - \corr_\mathcal{D}(\sign(H)) \leq \Phi'_\mathcal{D}(H, \sign(H)) - \Phi'_\mathcal{D}(H, h^*) $.
\end{lemma}
\begin{proof}
    We start by noting that 
    \begin{align*}
        \Phi'_\mathcal{D}(H, \sign(H)) - \Phi'_\mathcal{D}(H, h^*) &= \E_{(x,y)\sim \mathcal{D}} \left[(\psi'(H(x)) - y)(\sign(H(x))-h^*(x))\right]\\
        &= \E_{(x,y)\sim \mathcal{D}} \left[(1- y\psi'(H(x)))y(h^*(x)-\sign(H(x)))\right],
    \end{align*}
    where in the last line we use the fact that $y^2=1$.

    Consider any $(x,y)$ such that $yH(x)>0$: Here $y(h^*(x)-\textrm{sign}(H(x)))<0$. Furthermore, since $y$ and $H(x)$ have the same sign, so do $y$ and $\psi'(H(x))$ by \cref{lem:phimain}, and hence $(1- y\psi'(H(x)))\leq 1$. Similarly, whenever $yH(x)<0$: Then $y(h^*(x)-\textrm{sign}(H(x)))>0$, and $y$ and $\psi'(H(x))$ have opposite signs that imply $(1- y\psi'(H(x)))\geq 1$.

	Now the claim follows as
    \begin{align*}
        &\Phi'_\mathcal{D}(H, \sign(H)) - \Phi'_\mathcal{D}(H, h^*)) \\
        &= \E \left[ \mathds{1}_{yH(x)\geq 0}\underbrace{(1- y\psi'(H(x)))}_{\leq 1}\underbrace{y(h^*(x)-\sign(H(x)))}_{\leq 0} + \mathds{1}_{yH(x)< 0}\underbrace{(1- y\psi'(H(x)))}_{\geq 1}\underbrace{y(h^*(x)-\sign(H(x)))}_{\geq 0}\right]\\
        &\geq \E_{(x,y)\sim \mathcal{D}} [ y(h^*(x)-\sign H(x))]\\
        &= \corr_\mathcal{D}(h^*) - \corr_\mathcal{D}(\sign(H)).
    \end{align*}	
\end{proof}

We are now ready to prove the main result.
\begin{proof}[Proof of \Cref{thm:main}]
Let us dispense with the random events at once. The success of \Cref{lem:concmain}, the event that $\max_{t\in [T]} |\corr_{\widehat{D}_0} (\sign H_t) - \corr_{\mathcal{D}} (\sign H_t) |\leq\varepsilon/10$, and $\max_{t\in [T]} |\corr_{\widehat{D}'_t} (W_t) - \corr_{\mathcal{D}_t} (W_t) |\leq\gamma\varepsilon/10$ can be ensured with probability $1-\delta$ by a simple application of Hoeffing's inequality and union bound given the setting of $m$ and $S_0$ in the statement of the theorem. Similarly, $\varepsilon_{\textrm{Gen}}\leq\gamma\varepsilon/10$ holds in \Cref{lem:concmain} for $S,U = \Omega((\VCB+\log\delta^{-1})/\gamma^2\varepsilon^2)$. 

Let $h_t = (H_{t+1}-H_t)/\eta$. Since $\phi$ is $1$-smooth by \Cref{lem:phimain}:
\begin{align}\label{eq:smoothmain}
	\Phi_{\mathcal{D}}(H_{t+1}) - \Phi_\mathcal{D}(H_t) &\leq \E_{(x,y)\sim \mathcal{D}} \left[\eta \phi'(H_t(x),y) h_t(x) + \frac{\eta^2(h_t(x))^2}{2}\right] \nonumber \\
	&\leq \eta \Phi'_\mathcal{D}(H_t, h_t) + \frac{\eta^2}{2\gamma^2}.
\end{align}

{\bf Case A}: Consider any step $t$ where $\corr_{\widehat{D}'_t}(W_t) >\tau$. Here $h_t = W_t/\gamma$. It follows from \Cref{lem:concmain} that
\begin{align*}
	\Phi'_\mathcal{D}(H_t, h_t) &\leq -\frac{2\corr_{\mathcal{D}_t}(h_t)}{\gamma} + \frac{2\varepsilon_{\textrm{Gen}}}{\gamma}\\
	&\leq -\frac{2\corr_{\widehat{D}'_t}(h_t)}{\gamma} + \frac{2\varepsilon_{\textrm{Gen}}}{\gamma} +\frac{\varepsilon}{5} \\
    &\leq -\varepsilon
\end{align*}
where we set $\tau=\gamma\varepsilon$ and $\eta =\gamma^2\varepsilon$. By \Cref{eq:smoothmain}, the potential drops as $\Phi_\mathcal{D}(H_{t+1}) -\Phi_\mathcal{D}(H_t)\leq -\gamma^2\varepsilon^2/2$.

{\bf Case B}: Consider any step $t$ where $\corr_{\widehat{D}'_t}(W_t) \leq \tau$ and {\em crucially} $\Phi'_\mathcal{D}(H_t, \sign H_t) \geq \varepsilon$. Here $h_t = -\sign H_t$. Since $\Phi'_\mathcal{D}(H_t, h_t) = -\Phi'_\mathcal{D}(H_t, \sign H_t)\leq -\varepsilon$, by \Cref{eq:smoothmain}, we have $\Phi_\mathcal{D}(H_{t+1}) -\Phi_\mathcal{D}(H_t)\leq -\gamma^2\varepsilon^2/2$.

By \Cref{lem:phimain}, at initialization, $\Phi_{D}(\mathbf{0})$ is at most a half away from the minimum. Thus, setting $T=2/\gamma^2\varepsilon^2$, there must arise an iterate such that neither Case A nor Case B hold. That is, there is some $s\in [T]$ such that $\corr_{\widehat{D}'_s}(W_s) \leq \tau$ and $\Phi'_\mathcal{D}(H_s, \sign H_s) \leq \varepsilon$. Now using \Cref{lem:concmain} and that the weak learner $\gamma$-approximately maximizes correlation (\Cref{def:wl}), we have
\begin{align*}
	-\Phi'_\mathcal{D}(H_s, h^*) &\leq 2\corr_{\mathcal{D}_s}(h^*) + 2\varepsilon_{\textrm{Gen}} \\
	&\leq \frac{2\corr_{\mathcal{D}_s}(W_s)}{\gamma} + \frac{2\varepsilon_0}{\gamma} + 2\varepsilon_{\textrm{Gen}}\\
	&\leq \frac{2\corr_{\widehat{D}'_s}(W_s)}{\gamma} + \frac{2\varepsilon_0}{\gamma} + \frac{2\varepsilon}{5} \\
    &\leq \frac{2\varepsilon_0}{\gamma} + \frac{12\varepsilon}{5}
\end{align*}
where in the last line we recall $\tau=\varepsilon\gamma$ and $\varepsilon_{\textrm{Gen}}=\varepsilon\gamma/10$.

By \Cref{lem:consmain}, we have 
\begin{align*}
	\corr_\mathcal{D}(h^*) - \corr_\mathcal{D}(\sign(H_s)) & \leq \Phi'_\mathcal{D}(H_s, \sign(H_s)) - \Phi'_\mathcal{D}(H_s, h^*) \\
    &\leq \frac{2\varepsilon_0}{\gamma} + \frac{17\varepsilon}{5}.
\end{align*}

To complete the proof, we observe that
\begin{align*}
	\corr_\mathcal{D}(\overline{h}) &\leq \corr_{\widehat{D}_0}(\overline{h}) + \varepsilon/10\\
	&\leq \corr_{\widehat{D}_0}(\sign H_s) + \varepsilon/10\\
	&\leq \corr_{\mathcal{D}}(\sign H_s) + \varepsilon/5 \\
    &\leq 2\varepsilon_0/\gamma + 18\varepsilon/5,
\end{align*}
where we use the fact that $\overline{h}$ maximizes the empirical correlation on the dataset $\widehat{D}_0$. Substituting $\varepsilon$ appropriately yields the claim. 
\end{proof}

%% file: icml/variance.tex
\section{Improving unlabeled sample efficiency}
In this section, we reduce the number of unlabeled samples needed to $1/\gamma^3\varepsilon^3$, using the data reuse scheme from \citet{ghai2024sampleefficient}, which is crucially only applied to unlabeled data. The key idea behind the scheme is that since $H_t$ changes by a small amount each time, the change it induces on any twice-continuously differentiable potential is also small. Therefore, the desired relabeling distributions are not too different across rounds, and one may be able to reuse the distribution of past rounds to some extent. This difference is reflected in \Cref{alg:rev2} on Line 6 which permits the recursive use of unlabeled data from past rounds.
\begin{algorithm}[ht]
\begin{algorithmic}[1]
    \STATE \textbf{Inputs:} Samplers for labeled data from $\mathcal{D}$ and unlabeled data from $\mathcal{D}_\mathcal{X}$, $\gamma$-agnostic weak learning oracle $\mathcal{W}$, parameters $\eta, T, \tau, S, U, S_0, m, \sigma$.
    \STATE Initialize a zero hypothesis $H_1=\mathbf{0}$.
    \STATE Sample $S$ {\em labeled} examples to create dataset $\widehat{D}$.
    \FOR{$t=1$ to $T$}
        \STATE Sample $U$ {\em unlabeled} examples to create  dataset $\widehat{D}_t$.
        \STATE Construct a resampling distribution $\mathcal{D}'_t$ that picks $x$ uniformly from $\widehat{D}_t$, picks $\widehat{y}\in \{\pm 1\}$ uniformly and returns $(x,\widehat{y})$ if $t=1$; for $t>1$, do:
            \begin{enumerate}[label=\Alph*.,itemsep=-0.2em, topsep=0.2em]
            	\item With probability $1-\sigma$, return a sample $(x,\widehat{y})$ from $\mathcal{D}'_{t-1}$.
            	\item Else return $(x,\widehat{y})$ where $x$ is uniformly chosen from $\widehat{D}_t$, $\eta'\sim \textrm{Unif}[0,\eta]$, and $\widehat{y}$ is created as 
            $$\widehat{y} =\begin{cases}
            	+1 & \text{with probability } p_t(x,\eta'),\\
            	-1 & \text{with remaining probability, where}
            \end{cases}$$
            \end{enumerate}
            \vspace{-1em}
            \begin{align*}
            	p_t(x,\eta') &= \frac{1}{2}-\frac{\sigma\psi'(H_{t-1}(x))}{2(\eta+\sigma)} -\frac{\eta\psi''(H_{t-1}(x)+\eta' h_{t-1}(x)) h_{t-1}(x)}{2(\eta+\sigma)}.
            \end{align*}
        \STATE Construct a resampling distribution $\mathcal{D}_t$ that chooses between steps A and B with equal probability.
            \begin{enumerate}[label=\Alph*.,itemsep=-0.2em, topsep=0.2em]
            	\item Return $(x,y)$ picked uniformly from $\widehat{D}$.
            	\item Return $(x,\widehat{y})$ sampled uniformly from $\mathcal{D}'_t$.
            \end{enumerate}
        \STATE Sample $m$ times from $\mathcal{D}_t$ to create dataset $\widehat{D}'_t$.
        \STATE Call the weak learner on $\widehat{D}'_t$ to get $W_t=\mathcal{W}(\widehat{D}'_t)$.
        \IF{$\corr_{\widehat{D}'_t}(W_t) = \sum_{(x,\widehat{y})  \in \widehat{D}'_t} \widehat{y} W_t(x) >\tau$}
        	\STATE Update $H_{t+1} = H_t + \eta W_t/\gamma$.
        \ELSE
        	\STATE Update $H_{t+1} = H_t - \eta \sign(H_t)_t$.
        \ENDIF
    \ENDFOR
    \STATE Sample $S_0$ {\em labeled} examples to create dataset $\widehat{D}_0$.
    \STATE \textbf{Output} $\overline{h} = \argmax\limits_{h\in \{\sign(H_t):t\in [T]\}} \sum\limits_{(x,y)\in \widehat{D}_0} yh(x)$.
    \caption{Agnostic Boosting with Unlabeled Data Reuse} \label{alg:rev2}
\end{algorithmic}
\end{algorithm}
To do this, we first construct a twice-continuously differentiable potential using a Pseudo-Huber loss.
\begin{align}\label{eq:phi2}
\phi(z,y) = \psi(z)-yz, \text{ and } \psi(z) = \sqrt{1+x^2}-x.
\end{align}

\begin{theorem}[Main theorem with unlabeled data reuse]\label{thm:main2}
For any $\varepsilon,\delta>0$, there is an instantiation of parameters such that $\eta=\mathcal{O}(\gamma^2\varepsilon/\log |\mathcal{B}|)$, $T=\mathcal{O}(\log |\mathcal{B}|/\gamma^2\varepsilon^2)$, $\tau=\mathcal{O}(\gamma\varepsilon)$, $S=\mathcal{O}(\VCB/\gamma^2\varepsilon^2)$, $U=\mathcal{O}(1/\gamma\varepsilon)$, $S_0=\mathcal{O}(1/\varepsilon^2)$, $m=m(\varepsilon_0,\delta_0)+\mathcal{O}(1/\gamma^2\varepsilon^2)$ for which \Cref{alg:rev2} guarantees with probability $1-\delta - T\delta_0$ that \[\corr_\mathcal{D}(\overline{h}) \geq \max_{h\in \mathcal{H}} \corr_\mathcal{D}(h) - \frac{2\varepsilon_0}{\gamma} - \varepsilon.\]
During its execution, \Cref{alg:rev2} makes $T=\mathcal{O}(\log |\mathcal{B}|/\gamma^2\varepsilon^2)$ calls to the weak learner, and samples $S+S_0=\mathcal{O}(\log |\mathcal{B}|/\gamma^2\varepsilon^2)$ labeled samples and $TU=\mathcal{O}(\log |\mathcal{B}|/\gamma^3\varepsilon^3)$ unlabeled samples. 
\end{theorem}

Although the above result is always better in terms of the demand of unlabeled samples, it comes at the cost of increased oracle complexity, i.e., the number of calls to the weak learner, which now has a $\log |\mathcal{B}|$ factor unlike \Cref{thm:main}. Again mirroring techniques in \citet{ghai2024sampleefficient} provides some mitigation. In particular, in \Cref{sec:appvar2}, we provide a different guarantee for the same algorithm that makes $\mathcal{O}(1/\gamma^2\varepsilon^2)$ calls to the weak learner, while needing $\mathcal{O}(\log |\mathcal{B}|/\gamma^2\varepsilon^2)$ labeled and $\mathcal{O}(\log |\mathcal{B}|/\gamma^3\varepsilon^3 + (\log |\mathcal{B}|)^3/\gamma^2\varepsilon^2)$ unlabeled samples. These bounds can extend to infinite classes, as mentioned in \citet{ghai2024sampleefficient}, but we do not pursue it here.

%% file: icml/shift.tex
\section{Resiliency against covariate shift}
We consider the setting when the learner has access to a distribution $\mathcal{D}$ supported over $\mathcal{X}\times \{\pm 1\}$ and a different, possibly unrelated,  distribution $\mathcal{Q}$ supported over features $\mathcal{X}$. We show that under mild conditions the learner can still produce an arbitrarily accurate classifier. To measure how different $\mathcal{Q}$ and $\mathcal{D}_\mathcal{X}$ are, we define $C_\mathcal{X} = \left\|d\mathcal{D}_\mathcal{X}/d\mathcal{Q}\right\|_\infty$, which is a uniform upper bound on the Radon-Nikodym derivative of $\mathcal{D}_\mathcal{X}$ with respect to $\mathcal{Q}$.

\begin{theorem}[Main theorem for covariate shift]\label{thm:maincov}
For any $\varepsilon,\delta>0$, there is an algorithm that makes $\mathcal{O}(\mathcal{C}_\mathcal{X}/\gamma^2\varepsilon^2)$ calls to the weak learner, samples $\mathcal{O}(\VCB/\gamma^2\varepsilon^2)$ labeled samples from $\mathcal{D}$ and $TU=\mathcal{O}((\mathcal{C}_\mathcal{X})^3\VCB/\gamma^4\varepsilon^4)$ unlabeled samples from $\mathcal{Q}$, and outputs $\overline{h}$ such that with probability $1-\delta - T\delta_0$ that \[\corr_\mathcal{D}(\overline{h}) \geq \max_{h\in \mathcal{H}} \corr_\mathcal{D}(h) - \frac{(1+\mathcal{C}_\mathcal{X})\varepsilon_0}{\gamma} - \varepsilon.\] 
\end{theorem}

Recall that $\varepsilon_0$ can be made arbitrarily small by feeding more samples to the weak learner from the empirical distribution. Although this comes at a computational cost, in particular, the weak learner now needs to be more accurate, the sample complexity remains unaffected.

The key step is in \Cref{lem:consmain4}; it proves an analogue of \Cref{lem:consmain} implying that oversampling the part of the potential connected to the Huber loss does not hurt the correlation gap. From there, we set up a non-scalar potential measure to keep track of the progress of the learner. 

%% file: icml/applications.tex
\section{Applications}\label{sec:applications}
\subsection{Agnostic learning of halfspaces}
We illustrate how our method, when used as a black box, agnostically learns halfspaces over the $n$-dimensional Boolean hypercube under uniform marginals on the features.  Since unlabeled samples can be drawn from the uniform distribution at essentially no statistical cost, the added complexity of acquiring unlabeled data becomes solely a computational concern.

While this procedure does not match the best known upper bounds for halfspaces, it improves upon existing \emph{boosting}-based approaches.  In particular, building on \citet{kanade2009potential} and \citet{ghai2024sampleefficient}, we rely on empirical risk minimization (ERM) over all parities of degree at most \(d \approx 1/\varepsilon^4\).  Such an ERM rule achieves a weak learner advantage \(\gamma = n^{-d}\).  As shown in \cref{thm:half} below (proved in \cref{sec:proof_half}), this instantiation of our boosting framework reduces the labeled sample complexity from \(\mathcal{O}(\varepsilon^{-7} n^{60\varepsilon^{-4}})\) to \(\mathcal{O}(\varepsilon^{-6} n^{40\varepsilon^{-4}})\).

\begin{theorem}\label{thm:half}
  Let $\mathcal{D}$ be any distribution over \(\{\pm 1\}^n \times \{\pm 1\}\) with uniform feature marginals, and let
  \[
    \mathcal{H} \;=\; \bigl\{\sign\bigl(w^\top x \;-\;\theta\bigr) \;\mid\; (w,\theta)\in\mathbb{R}^{n+1}\bigr\}
  \]
  denote the class of halfspaces.
  There exists a degree \(d = O(\varepsilon^{-4})\) such that running \cref{alg:rev} with ERM over parities of degree at most \(d\) produces a classifier \(\overline{h}\) satisfying
  \[
    l_{\mathcal{D}}\bigl(\overline{h}\bigr) 
    \;\le\; 
    \min_{h \in \mathcal{H}}\, l_{\mathcal{D}}(h)\;+\;\varepsilon,
  \]
  while using only \(\mathcal{O}\!\bigl(\varepsilon^{-6}\,n^{40\,\varepsilon^{-4}}\bigr)\) labeled samples in \(n^{\mathrm{poly}(1/\varepsilon)}\) time.
\end{theorem}

\subsection{Boosting for reinforcement learning}\label{sec:rl}
The construction of near-optimal policy for reinforcement learning (RL) via boosting was first pursued in \citet{brukhim2022boosting}. \citet{ghai2024sampleefficient} improve on these results. Modifying the RL setting to include the ability to sample trajectories \emph{without observing reward}, we can apply our results to reduce the number of samples that require reward feedback. Such a feedback model could be useful where rollouts are cheap but reward feedback is not because it comes from human label or expensive processes \citep{finn2016generalizing}. In \cref{app:rl} we provide a formal description of this modified setting.

Plugging \Cref{alg:rev} into a modified meta-algorithm of \citet{brukhim2022boosting} which allows for trajectories without reward yields the following result for binary-action MDPs. Our result improves upon \citet{ghai2024sampleefficient} in that it requires fewer with-reward episodes. Here, $V^\pi$ is the expected discounted reward of a policy $\pi$, $V^*$ is its maximum. $\beta$ is the discount factor of the underlying MDP, and $C_\infty,D_\infty$ and $\mathcal{E}, \mathcal{E}_\nu$ are distribution mismatch and policy completeness terms (related to the inherent Bellman error). In the {\em episodic model}, the learner interacts with the MDP in episodes. In the {\em $\nu$-reset model}, the learner can seed the initial state with a fixed well dispersed distribution $\nu$ as a means to exploration. See \Cref{app:rl} for a complete statement of results and details of the setting. 

\begin{theorem}[Informal; stated formally in \Cref{thm:rl2}]\label{thm:rl}
Let $\mathcal{W}$ be a $\gamma$-weak learner for the policy class $\Pi$ operating with a base class $\mathcal{B}$, with sample complexity $m(\varepsilon_0, \delta_0) = (\log |\mathcal{B}|/\delta_0)/\varepsilon_0^2$. Fix tolerance $\varepsilon$ and failure probability $\delta$. In the {\em episodic} access model, there is an algorithm using that uses the weak learner $\mathcal{W}$ to produce a policy $\overline{\pi}$ such that with probability $1-\delta$, we have
$ V^* - V^{\overline{\pi}} \leq {(C_\infty \mathcal{E})}/{(1-\beta)} + \varepsilon,$
while sampling $\mathcal{O}((\log |\mathcal{B}|)/\gamma^3 \varepsilon^4)$ episodes of length $\mathcal{O}((1-\beta)^{-1})$ without reward feedback and  $\mathcal{O}((\log |\mathcal{B}|)/\gamma^2 \varepsilon^3)$ episodes of length $\mathcal{O}((1-\beta)^{-1})$ with reward feedback. In the {\em $\nu$-reset} access model, there is a setting of parameters such that \Cref{alg:rlMAIN1} when given access to $\mathcal{W}$ produces a policy $\overline{\pi}$ such that with probability $1-\delta$, we have
$ V^* - V^{\overline{\pi}} \leq {(D_\infty \mathcal{E}_\nu)}/{(1-\beta)^2} + \varepsilon$, while sampling $\mathcal{O}((\log |\mathcal{B}|)/\gamma^3 \varepsilon^5)$ episodes of length $\mathcal{O}((1-\beta)^{-1})$ without reward feedback and $\mathcal{O}((\log |\mathcal{B}|)/\gamma^2 \varepsilon^4)$ episodes of length $\mathcal{O}((1-\beta)^{-1})$ with reward feedback.
\end{theorem}

%% file: icml/experiments.tex
\section{Experiments}

In this section, we demonstrate the empirical viability of our approach. \Cref{tab:boosting_results_updated} showcases the results from our initial experiments comparing \cref{alg:rev} with the agnostic boosting method introduced by \citet{kanade2009potential}, herein referred to as the Potential-based Agnostic Booster (PAB). These evaluations were performed on various UCI classification datasets \citep{misc_ionosphere_52,misc_spambase_94,smith1988using, statlog_(german_credit_data)_144, sonar, waveform}, employing decision stumps \citep{scikit} as the weak learners. Notably, \cref{alg:rev} extends PAB to handle unlabeled data by incorporating our newly defined potential function \eqref{eq:new_potential}. 

\begin{table*}[ht]
\centering
\scriptsize
\begin{tabular}{|c|c|c|c|c|c|c|c|c|}
\hline
\multicolumn{1}{|c|}{\textbf{Dataset}} & 
\multicolumn{2}{c|}{\textbf{No Added Noise}} & 
\multicolumn{2}{c|}{\textbf{5\% Noise}} & 
\multicolumn{2}{c|}{\textbf{10\% Noise}} & 
\multicolumn{2}{c|}{\textbf{20\% Noise}} \\
\hline
 & \textbf{PAB} & \textbf{Ours} & \textbf{PAB} & \textbf{Ours} & \textbf{PAB} & \textbf{Ours} & \textbf{PAB} & \textbf{Ours} \\
\hline
Ionosphere & 0.87 $\pm$ 0.05 & \textbf{0.91 $\pm$ 0.04} & 0.88 $\pm$ 0.05 & \textbf{0.90 $\pm$ 0.04} & 0.84 $\pm$ 0.06 & \textbf{0.90 $\pm$ 0.04} & 0.81 $\pm$ 0.06 & \textbf{0.83 $\pm$ 0.06} \\
\hline
Diabetes & 0.84 $\pm$ 0.09 & \textbf{0.89 $\pm$ 0.07} & \textbf{0.86 $\pm$ 0.08} & \textbf{0.86 $\pm$ 0.09} & 0.79 $\pm$ 0.09 & 0.79 $\pm$ 0.10 & 0.76 $\pm$ 0.10 & 0.80 $\pm$ 0.10 \\
\hline
Spambase & 0.91 $\pm$ 0.02 & \textbf{0.94 $\pm$ 0.02} & 0.90 $\pm$ 0.03 & \textbf{0.92 $\pm$ 0.03} & 0.89 $\pm$ 0.03 & \textbf{0.90 $\pm$ 0.02} & 0.83 $\pm$ 0.04 & \textbf{0.87 $\pm$ 0.03} \\
\hline
German & 0.79 $\pm$ 0.07 & \textbf{0.86 $\pm$ 0.07} & \textbf{0.84 $\pm$ 0.08} & \textbf{0.84 $\pm$ 0.08} & 0.76 $\pm$ 0.08 & \textbf{0.87 $\pm$ 0.07} & 0.75 $\pm$ 0.08 & \textbf{0.77 $\pm$ 0.08} \\
\hline
Sonar & 0.78 $\pm$ 0.08 & \textbf{0.92 $\pm$ 0.05} & 0.68 $\pm$ 0.09 & \textbf{0.89 $\pm$ 0.06} & 0.84 $\pm$ 0.08 & \textbf{0.87 $\pm$ 0.07} & 0.69 $\pm$ 0.10 & \textbf{0.77 $\pm$ 0.08} \\
\hline
Waveform & 0.89 $\pm$ 0.02 & \textbf{0.89 $\pm$ 0.02} & \textbf{0.88 $\pm$ 0.03} & 0.87 $\pm$ 0.03 & \textbf{0.86 $\pm$ 0.03} & 0.86 $\pm$ 0.03 & 0.83 $\pm$ 0.03 & \textbf{0.83 $\pm$ 0.03} \\
\hline
Average & 0.84 & \textbf{0.89} & 0.84 & \textbf{0.88} & 0.81 & \textbf{0.84} & 0.78 & \textbf{0.81} \\
\hline
\end{tabular}
\caption{50-fold  cross-validated accuracies of the Potential based Agnostic Booster (PAB) \citep{kanade2009potential} and our proposed boosting algorithm on six datasets with 0\%, 5\%, 10\%, and 20\% added label noise (during training). Sonar and Ionosphere have 50\% of labels dropped while the remaining datasets have 90\% of labels dropped. A final row is included for the average accuracy (evenly weighted) over all $6$ datasets.}
\label{tab:boosting_results_updated}
\end{table*}

To evaluate the algorithms' robustness against label noise, we introduced noise levels of 5\%, 10\%, and 20\% during the training phase. We randomly remove a certain percentage\footnote{Specifically, we omitted 50\% of labels for smaller datasets (Sonar and Ionosphere) and 90\% for the other datasets.} of labels from each dataset to create scenarios with both labeled and unlabeled instances. Our findings indicate that incorporating unlabeled examples leads to improved performance. This enhancement is likely attributed to the limitation of PAB in reusing samples, which consequently restricts the number of boosting iterations when the sample size is constrained. For a comprehensive overview of the experimental setup, please refer to \cref{sec:experiment_app}.

\section{Conclusion}
This paper aims to levereage unlabeled data to bring down the sample complexity of agnostic boosting. The theoretical improvements are stark. When given as much {\em unlabeled} data as the amount of {\em labeled} data required for existing approaches, the resultant sample complexity precipitates down to that of ERM. This is accomplished by a novel decomposable potential function, whose derivative naturally splits into two parts estimable by labeled and unlabeled data, respectively.

%% file: icml/app-var.tex
{\bf Map of the appendix.} In \Cref{sec:app-unlab}, we complete the proofs for results concerning improved unlabeled sample efficiency. \Cref{sec:app-cov} discusses the proofs for covariate shift. In \Cref{sec:experiment_app} are provided experimental details not found in the main paper. \Cref{sec:proof_half} and \Cref{app:rl} provide further details on applications of boosting to learning halfspace and reinforcement learning, respectively.

\section{Improving unlabeled sample efficiency}\label{sec:app-unlab}
\subsection{Proofs for the result}
Notice that in this section we use different choices of $\phi$ and $\psi$, those stated in \Cref{eq:phi2}. However, \Cref{lem:consmain} continues to hold. In fact, the latter only requires that $z\psi'(z)\geq 0$ for all $z$. To maintain the continuity of presentation, our organization and notation closely mirror \citet{ghai2024sampleefficient}. Throughout this section, we will always set $\sigma = \eta/\gamma$.

Define $\Psi_\mathcal{D}(H) = \E_{(x,y)\sim \mathcal{D}}[\Psi(H(x))]$ and $\Psi'_\mathcal{D}(H,h) = \E_{(x,y)\sim \mathcal{D}}[\Psi'(H(x))h(x)]$.
\begin{proof}[Proof of \Cref{thm:main2}.]
This proof is almost identical to that of Theorem 4 in \citet{ghai2024sampleefficient}. We reproduce it for completeness. In fact, the only change stems from the new upper bound on $\varepsilon_{\textrm{Gen}}$ in \Cref{lem:concmain2}, which unlike the previous work makes a distinction between labeled and unlabeled samples.

\begin{lemma}\label{lem:concmain2}
	There exists a universal constant $C>0$ such that with probability $1-\delta$, for all $t\in [T], h\in \mathcal{B}\cup \{h^*\}:$
	$$\left\lvert\Phi'_{\mathcal{D}}(H_t, h) + 3\corr_{\mathcal{D}_t}(h)\right\rvert \leq \underbrace{C\left(\sqrt{\frac{\log |\mathcal{B}| + \log \frac{1}{\delta}}{S}} + \frac{\eta}{\gamma}\left(\sqrt{\frac{\log |\mathcal{B}|T/\delta}{\sigma U}}+ \log |\mathcal{B}|T/\delta\right)\right)}_{\varepsilon_{\textrm{Gen}}}.$$
\end{lemma}

But before that let us dispense with the random events at once. The success of \Cref{lem:concmain2}, the event that $\max_{t\in [T]} |\corr_{\widehat{D}_0} (\sign H_t) - \corr_{\mathcal{D}} (\sign H_t) |\leq\varepsilon''/10$, and $\max_{t\in [T]} |\corr_{\widehat{D}'_t} (W_t) - \corr_{\mathcal{D}_t} (W_t) |\leq\varepsilon'/10$ can be ensured with probability $1-\delta$ by a simple application of Hoeffing's inequality and union bound given the setting of $m$ and $S_0$ appropriately. We will soon set precise values of $\varepsilon'$ and $\varepsilon''$.

Recall that $h_t = \eta(H_{t+1}-H_t)$. \Cref{eq:smoothmain} can be rearranged to get
	\begin{align*}
		-\frac{1}{T}\sum_{t=1}^T\Phi'_\mathcal{D}(H_t, h_t) &\leq \frac{\sum_{t=1}^T(\Phi_\mathcal{D}(H_t) - \Phi_\mathcal{D}(H_{t+1}))}{\eta T}  + \frac{\eta}{2\gamma^2} \leq \frac{2}{\eta T} + \frac{\eta}{2\gamma^2} 
	\end{align*}
	where we use the fact that $\phi(0,y)-\min_z\phi(z,y) \leq 1$.
 
{\bf Case A}: If $h_t=W_t/\gamma$, observe that $\text{corr}_{\mathcal{D}_t}(W_t) \geq \text{corr}_{{D}'_t}(W_t)-\varepsilon'/10 \geq \tau -\varepsilon'/10$. Now apply \Cref{lem:concmain2} to get
\begin{align*}
	-\Phi'_\mathcal{D}\left(H_t, h_t\right) \geq \frac{3}{\gamma}\text{corr}_{\mathcal{D}_t} (W_t) - \frac{\varepsilon_{\text{Gen}}}{\gamma} \leq \frac{3}{\gamma}\left(\tau-\frac{\varepsilon'}{10}\right) -\frac{\varepsilon_{\text{Gen}}}{\gamma}
\end{align*}

{\bf Case B}: If $h_t=-\text{sign}(H_t)$, then $\text{corr}_{\mathcal{D}_t}(W_t) \leq \text{corr}_{{D}'_t}(W_t)+\varepsilon'/10 \leq \tau +\varepsilon'/10$. Applying \Cref{lem:concmain2}, we get 
\begin{align*}
	3\left(\tau + \frac{\varepsilon'}{10}\right) &\geq 3\text{corr}_{\mathcal{D}_t} (W_t) \geq 3\gamma \text{corr}_{\mathcal{D}_t} (h^*) -3\varepsilon_0 \geq -\gamma \Phi'_\mathcal{D}(H_t, h^*) -3\varepsilon_0 - \gamma \varepsilon_{\text{Gen}}
\end{align*}
Using \Cref{lem:consmain}, this translates to 
\begin{align*}
\Phi'_\mathcal{D}(H_t,-\text{sign}(H_t)) &= -\Phi'_\mathcal{D}(H_t,\text{sign}(H_t)) \leq -	\Phi'_\mathcal{D}(H_t, h^*) -  (\text{corr}_\mathcal{D}(h^*)-\text{corr}_\mathcal{D}(\text{sign}(H_t)))\nonumber \\
&\leq  -(\text{corr}_\mathcal{D}(h^*)-\text{corr}_\mathcal{D}(\text{sign}(H_t))) + \frac{3}{\gamma}\left(\tau+\frac{\varepsilon'}{10}+\varepsilon_0\right) + \varepsilon_{\text{Gen}}.
\end{align*}

In either case, we have
\begin{align*}
-\Phi'_\mathcal{D} ({H}_t, h_t ) \geq \min\bigg\{& \frac{3}{\gamma}\left(\tau-\frac{\varepsilon'}{10}\right) - \frac{\varepsilon_{\text{Gen}}}{\gamma} ,  (\text{corr}_\mathcal{D}(h^*)-\text{corr}_\mathcal{D}(\text{sign}(H_t))) - \frac{3}{\gamma} \left(\tau+\frac{\varepsilon'}{10}+\varepsilon_0\right) - \varepsilon_{\text{Gen}} \bigg\}.
\end{align*}
Now, set $$\tau = \frac{1}{3}\left(\frac{{4}}{\eta T} + \frac{\eta}{\gamma^2}+\frac{\varepsilon_{\text{Gen}}}{\gamma}\right) \gamma+ \frac{\varepsilon'}{10}.$$ 
Hereafter let $s$ be the time step satisfying
\begin{align}
	 \text{corr}_\mathcal{D}(h^*)-\text{corr}_\mathcal{D}(\text{sign}(H_s)) &\leq \frac{3}{\gamma} (2\tau+\varepsilon_0) + \left(1-\frac{1}{\gamma}\right)\varepsilon_{\text{Gen}} \nonumber = \frac{8}{\eta T} + \frac{2\eta}{\gamma^2} + \left(1+\frac{1}{\gamma}\right)\varepsilon_\text{Gen} +3\left(\frac{\varepsilon_0}{\gamma} + \frac{\varepsilon'}{5\gamma}\right).
\end{align}
Such a choice must exists, since otherwise we get for all $t$ that $$-\Phi'_\mathcal{D} ({H}_t, h_t ) \geq \frac{3}{\gamma}\left(\tau-\frac{\varepsilon'}{10}\right) -\frac{\varepsilon_{\text{Gen}}}{\gamma} = \frac{{4}}{(\eta T)} + \frac{\eta}{\gamma^2},$$
which contradicts \Cref{eq:smoothmain}. 
Combining this with the observation that $\overline{h}$ minimizes the correlation on $\widehat{D}_0$, we get
\begin{align*}
	 \text{corr}_\mathcal{D}(h^*)-\text{corr}_\mathcal{D}(\overline{h})\leq  \frac{8}{\eta T} + \frac{2\eta}{\gamma^2} + \frac{2\varepsilon_\text{Gen}}{\gamma} +\frac{3}{\gamma} \left(\varepsilon_0 + \frac{\varepsilon'}{5}\right) + \frac{\varepsilon''}{5}.
\end{align*}
Setting $\varepsilon' = \gamma\varepsilon/100$, $\varepsilon'' = \gamma\varepsilon/100$ and plugging in the proposed hyper-parameters with appropriate constants yields the claimed result.
\end{proof}

\begin{proof}[Proof of \Cref{lem:concmain2}]
By the definition of $\mathcal{D}_t$, we have that 
\begin{align*}
\corr_{\mathcal{D}_t}(h) = \frac{1}{3}\widehat{\E}_{\widehat{D}} [yh(x)] + \frac{2}{3}{\E}_{\mathcal{D}'_t} [yh(x)] = \frac{1}{3}\widehat{\E}_{\widehat{D}} [yh(x)] + \frac{2}{3}\corr_{\mathcal{D}'_t}(h),
\end{align*}
Since $\widehat{D}$ is composed of IID draws from $\mathcal{D}$, the standard uniform convergence result via union bound gets that for some constant $C\geq 0$ we have with probability $1-\delta$ that $|\widehat{\E}_{\widehat{D}} [yh(x)] - \corr_\mathcal{D}(h)|$ is at most $\sqrt{(\log |\mathcal{B}|+\log \delta^{-1})/S}$.

By inspecting Line 6 in \Cref{alg:rev2} here and Line 5 in Algorithm 1 in \citet{ghai2024sampleefficient} with the substitution that $y=1$, we note that the two are identical. Therefore, we can apply the following result from \citet{ghai2024sampleefficient}.
\begin{lemma}[Lemma 6 in \citet{ghai2024sampleefficient}]
	Setting $\sigma=\eta/\gamma$. There exists a universal constant $C>0$ such that with probability $1-\delta$, for all $t\in [T], h\in \mathcal{B}\cup \{h^*\}:$
	$$\left\lvert\Psi'_{\mathcal{D}}(H_t, h) + 2\corr_{\mathcal{D}'_t}(h)\right\rvert \leq {\frac{C\eta}{\gamma}\left(\sqrt{\frac{\log |\mathcal{B}|T/\delta}{\sigma U}}+ \log |\mathcal{B}|T/\delta\right)}.$$
\end{lemma}

Since $\Phi_\mathcal{D}(H)=\Psi_{\mathcal{D}}(H) - \corr_\mathcal{D}(H)$, we get 
	\begin{align*}
		\left\lvert\frac{1}{3}\Phi'_{\mathcal{D}}(H_t, h) + \corr_{\mathcal{D}_t}(h)\right\rvert \leq \frac{1}{3} |\widehat{\E}_{\widehat{D}} [yh(x)] - \corr_\mathcal{D}(h)| + \frac{1}{3}\left\lvert\Psi'_{\mathcal{D}}(H_t, h) + 2\corr_{\mathcal{D}'_t}(h)\right\rvert 
	\end{align*}
\[ \Phi'_\mathcal{D}(H_t,h) = \E_{x\sim \mathcal{D}_\mathcal{X}} [\psi'(H_t(x))h(x)] - \E_{(x,y)\sim \mathcal{D}} [yh(x)], \]
completing the proof. 
\end{proof}

\subsection{Trading off oracle Complexity and unlabeled sample complexity}\label{sec:appvar2}

\begin{theorem}[Main theorem with unlabeled data reuse]\label{thm:main3}
For any $\varepsilon,\delta>0$, there is an instantiation of parameters for which \Cref{alg:rev2} guarantees with probability $1-\delta - T\delta_0$ that \[\corr_\mathcal{D}(\overline{h}) \geq \max_{h\in \mathcal{H}} \corr_\mathcal{D}(h) - \frac{2\varepsilon_0}{\gamma} - \varepsilon.\]
During its execution, \Cref{alg:rev2} makes $\mathcal{O}(1/\gamma^2\varepsilon^2)$ calls to the weak learner, and samples $S+S_0=\mathcal{O}(\log |\mathcal{B}|/\gamma^2\varepsilon^2)$ labeled samples and $TU=\mathcal{O}(\log |\mathcal{B}|/\gamma^3\varepsilon^3+(\log |\mathcal{B}|)^3/\gamma^3\varepsilon^2)$ unlabeled samples. 
\end{theorem}

\begin{proof}
The proof is identical to that of \Cref{thm:main2}, except crucially to the substitution of the following bound on $\varepsilon_{\textrm{Gen}}$.

There exists a universal constant $C>0$ such that with probability $1-\delta$, for any $t\in [T], h\in \mathcal{B}\cup \{h^*\}:$
	$$\left\lvert\frac{1}{3}\Phi'_\mathcal{D}(H_t, h) + \corr_{\mathcal{D}_t}(h)\right\rvert \leq \underbrace{C\left(\sqrt{\frac{\log |\mathcal{B}| + \log \frac{1}{\delta}}{S}} + \left(\sigma + \frac{\eta}{\gamma}\right)\left(\sqrt{\frac{\log |\mathcal{B}|T/\delta}{\sigma U}}+ {\frac{(\log |\mathcal{B}|T/\delta)^{3/2}}{\sqrt{U}}}\right)\right)}_{\varepsilon_{\textrm{Gen}}}.$$

To prove this claim itself, we follow the same recipe as in the proof of \Cref{lem:concmain2}. Once again we observe that By Line 6 in \Cref{alg:rev2} here is identical to Line 5 in Algorithm 1 in \citet{ghai2024sampleefficient} with the substitution that $y=1$. Therefore, we can apply the following result from \citet{ghai2024sampleefficient}.
\begin{lemma}[Lemma 15 in \citet{ghai2024sampleefficient}]
	Setting $\sigma=\eta/\gamma$. There exists a universal constant $C>0$ such that with probability $1-\delta$, for all $t\in [T], h\in \mathcal{B}\cup \{h^*\}:$
	$$\left\lvert\Psi'_{\mathcal{D}}(H_t, h) + 2\corr_{\mathcal{D}'_t}(h)\right\rvert \leq {\frac{C\eta}{\gamma}\left(\sqrt{\frac{\log |\mathcal{B}|T/\delta}{\sigma U}}+ \log |\mathcal{B}|T/\delta\right)}.$$
\end{lemma}	
\end{proof}

\section{Resiliency against covariate shift}\label{sec:app-cov}
Let $\psi(z)$ be the Huber loss. Instead of definite a scalar potential, this time we will directly define the population potential measure that involves both $\mathcal{D}$ and $\mathcal{Q}$. Recall that $\mathcal{C}_\mathcal{X} \geq \|d\mathcal{D}_\mathcal{X}/d\mathcal{Q}\|$ has to be at least one.
\begin{align*}
    \Phi(H) &= \mathcal{C}_\mathcal{X}\E_{x\sim \mathcal{Q}} [\psi(H(x))] - \E_{(x,y)\sim \mathcal{D}}[yH(x)]\\
    \Phi'(H,h) &= \mathcal{C}_\mathcal{X}\E_{x\sim \mathcal{Q}} [\psi'(H(x))h(x)] - \E_{(x,y)\sim \mathcal{D}}[yh(x)]
\end{align*}

We describe a variant of \Cref{alg:rev} that can tolerate a mismatch between $\mathcal{D}_\mathcal{X}$ and $\mathcal{Q}$. The key modification happens in Line 6 of \Cref{alg:rev4}, where pseudo-labeled samples created from the unlabeled distribution $\mathcal{Q}$ are sampled a higher rate than labeled samples.

\begin{algorithm}[ht]
\begin{algorithmic}[1]
    \STATE \textbf{Inputs:} Samplers for labeled data from $\mathcal{D}$ and unlabeled data from $\mathcal{Q}$, $\gamma$-agnostic weak learning oracle $\mathcal{W}$, parameters $\eta, T, \tau, S, U, S_0, m$.
    \STATE Initialize a zero hypothesis $H_1=\mathbf{0}$.
    \STATE Sample $S$ {\em labeled} examples to create dataset $\widehat{D}$.
    \FOR{$t=1$ to $T$}
        \STATE Sample $U$ {\em unlabeled} examples to create  dataset $\widehat{D}_t$.
        \STATE Construct a resampling distribution $\mathcal{D}_t$ that:
            \begin{enumerate}[label=\Alph*.,itemsep=-0.2em, topsep=0.2em]
            	\item With probability $\frac{1}{\mathcal{C}_\mathcal{X}}$, returns $(x,y)$ picked uniformly from $\widehat{D}$.
            	\item With remaining probability, picks $x$ uniformly from $\widehat{D}_t$, and returns $(x, \widehat{y})$, where $\widehat{y}$ is chosen as 
            \end{enumerate}
            $$\widehat{y} =\begin{cases}
            	+1 & \text{with probability } p_t(x) = \frac{1-\psi'(H_t(x))}{2},\\
            	-1 & \text{with remaining probability.}
            \end{cases}$$
        \STATE Sample $m$ times from $\mathcal{D}_t$ to create dataset $\widehat{D}'_t$.
        \STATE Call the weak learner on $\widehat{D}'_t$ to get $W_t=\mathcal{W}(\widehat{D}'_t)$.
        \IF{$\corr_{\widehat{D}'_t}(W_t) = \sum_{(x,\widehat{y})  \in \widehat{D}'_t} \widehat{y} W_t(x) >\tau$}
        	\STATE Update $H_{t+1} = H_t + \eta W_t/\gamma$.
        \ELSE
        	\STATE Update $H_{t+1} = H_t - \eta \sign(H_t)_t$.
        \ENDIF
    \ENDFOR
    \STATE Sample $S_0$ {\em labeled} examples to create dataset $\widehat{D}_0$.
    \STATE \textbf{Output} $\overline{h} = \argmax\limits_{h\in \{\sign(H_t):t\in [T]\}} \sum\limits_{(x,y)\in \widehat{D}_0} yh(x)$.
    \caption{Covariate-shift Resistant Agnostic Boosting with Unlabeled Data} \label{alg:rev4}
\end{algorithmic}
\end{algorithm}

The uniform convergence still applies after some minor adjustment.
\begin{lemma}\label{lem:concmain4}
	There exists a universal constant $C>0$ such that with probability $1-\delta$, for all $t\in [T], h\in \mathcal{B}\cup \{h^*\}$,
	$$\left\lvert\Phi'(H_t, h) + (1+\mathcal{C}_\mathcal{X})\corr_{\mathcal{D}_t}(h)\right\rvert \leq \varepsilon_{\textrm{Gen}}\coloneqq C \left(\sqrt{\frac{\VCB + \log \frac{1}{\delta}}{S}} + \mathcal{C}_\mathcal{X} \sqrt{\frac{\VCB + \log \frac{1}{\delta}}{U}}\right).$$
\end{lemma}
\begin{proof}[Proof of \Cref{lem:concmain4}]
By the definition of $\mathcal{D}_t$, we have that 
\begin{align*}
\corr_{\mathcal{D}_t}(h) = \frac{1}{1+\mathcal{C}_\mathcal{X}}\left(\widehat{\E}_{\widehat{D}} [yh(x)] - \mathcal{C}_\mathcal{X}\widehat{\E}_{\widehat{D}_t} [\psi'(H_t(x))h(x)]\right),
\end{align*}
where we use the fact that Line 6.B in \Cref{alg:rev4} ensures $\E[\widehat{y}|x] = -\psi'(H_t(x))$. Since $\widehat{D}$ and $\widehat{D}_t$ are composed of IID draws from $\mathcal{D}$ and $\mathcal{Q}$ respectively, we have that, for some constant $C\geq 0$ we have with probability $1-\delta$ that $|\widehat{\E}_{\widehat{D}} [yh(x)] - \corr_\mathcal{D}(h)|\leq \sqrt{({\VCB+\log \delta^{-1})}/{S}}$ and $|\widehat{\E}_{\widehat{D}_t} [\psi'(H_t(x))h(x)] - \E_{x\sim \mathcal{Q}}[\psi'(H_t(x))h(x)] | \leq \sqrt{({\VCB+\log \delta^{-1})}/{U}}$. 
\begin{align*}
	|(1+\mathcal{C}_\mathcal{X}) \corr_{\mathcal{D}_t}(h) + \Phi'(H_t, h)| \leq \mathcal{C}_\mathcal{X}|\widehat{\E}_{\widehat{D}_t} [\psi'(H_t(x))h(x)] - \E_{x\sim \mathcal{Q}}[\psi'(H_t(x))h(x)] | + |\widehat{\E}_{\widehat{D}} [yh(x)] - \corr_\mathcal{D}(h)|
\end{align*}
The decomposition above completes the proof.
\end{proof}

We will need the following well-known property of Random-Nikodym derivatives.

\begin{lemma}\label{lem:shiftdist}
    For any non-negative function $f:\mathcal{X}\to \mathbb{R}_{\geq 0}$, it is true that $ C_\mathcal{X}\E_{x\sim Q}[f(x)] \geq \E_{x\sim \mathcal{D}_\mathcal{X}} [f(x)]$.
\end{lemma}
\begin{proof}
Since $\mathcal{C}_\mathcal{X} \geq \|d\mathcal{D}_\mathcal{X}/d\mathcal{Q}\|$, we have $C_\mathcal{X}\E_{x\sim Q}[f(x)] \geq \E_{x\sim Q}\left[f(x)\frac{d\mathcal D{x}}{d\mathcal{Q}}(x)\right] = \E_{x\sim \mathcal{D}_\mathcal{X}} [f(x)]$.
\end{proof}

The key idea in the analysis occurs in the following lemma. Essentially, it says oversampling the $\psi$ part from unlabeled data does not hurt the correlation gap.
\begin{lemma}\label{lem:consmain4}
    For any real-valued classifier $H:\mathcal{X}\to \mathbb{R}$, we have $$\Phi'(H, \sign(H)) - \Phi'(H, h^*) \geq \corr_\mathcal{D}(h^*) - \corr_\mathcal{D}(\sign(H)).$$
\end{lemma}
\begin{proof}
Using \Cref{lem:shiftdist} below, we arrive at 
    \begin{align*}
        \Phi'(H, \sign(H)) - \Phi'(H, h^*) &= \mathcal{C}_\mathcal{X}\E_{x\sim \mathcal{Q}} \left[\psi'(H(x))(\sign(H(x))-h^*(x))\right] - \E_{(x,y)\sim \mathcal{D}} \left[y(\sign(H(x))-h^*(x))\right]\\
        &\geq \E_{x\sim \mathcal{D}_\mathcal{X}} \left[\psi'(H(x))(\sign(H(x))-h^*(x))\right] - \E_{(x,y)\sim \mathcal{D}} \left[y(\sign(H(x))-h^*(x))\right]\\
        &= \E_{(x,y)\sim \mathcal{D}} \left[(\psi'(H(x))-y)(\sign(H(x))-h^*(x))\right].
    \end{align*}
Recall that $z$ and $\psi'(z)$ always have the same sign, and hence so do $\psi'(z)$ and $\sign(z)$. This ensures non-negativity as $\psi'(H(x))(\sign(H(x))-h^*(x)) = |\psi'(H(x))|-\psi'(H(x))h^*(x)$, since $h^*(x)$ is restricted to $\{\pm 1\}$.

From here onward, our original proof strategy work. In particular since $y^2=1$, we get
    \begin{align*}
        \Phi'(H, \sign(H)) - \Phi'(H, h^*) \geq \E_{(x,y)\sim \mathcal{D}} \left[(1-y\psi'(H(x)))y(\sign(H(x))-h^*(x))\right].
    \end{align*}
As before, consider any $(x,y)$ such that $yH(x)>0$: Here $y(h^*(x)-\textrm{sign}(H(x)))<0$. Furthermore, since $y$ and $H(x)$ have the same sign, so do $y$ and $\psi'(H(x))$, and hence $(1- y\psi'(H(x)))\leq 1$. Similarly, whenever $yH(x)<0$: Then $y(h^*(x)-\textrm{sign}(H(x)))>0$, and $y$ and $\psi'(H(x))$ have opposite signs that imply $(1- y\psi'(H(x)))\geq 1$.

	Now the claim follows as
    \begin{align*}
        &\Phi'_\mathcal{D}(H, \sign(H)) - \Phi'_\mathcal{D}(H, h^*)) \\
        = &\E_{(x,y)\sim \mathcal{D}_\mathcal{X}} \left[ \mathds{1}_{yH(x)\geq 0}\underbrace{(1- y\psi'(H(x)))}_{\leq 1}\underbrace{y(h^*(x)-\sign(H(x)))}_{\leq 0}+ \mathds{1}_{yH(x)< 0}\underbrace{(1- y\psi'(H(x)))}_{\geq 1}\underbrace{y(h^*(x)-\sign(H(x)))}_{\geq 0}\right]\\
        \geq&  \E_{(x,y)\sim \mathcal{D}} [ y(h^*(x)-\sign H(x))]\\
        =&  \corr_\mathcal{D}(h^*) - \corr_\mathcal{D}(\sign(H)).
    \end{align*}	
\end{proof}

We are finally ready to prove the main result.
\begin{proof}[Proof of \Cref{thm:main}]
Let us dispense with the random events at once. The success of \Cref{lem:concmain4}, the event that $\max_{t\in [T]} |\corr_{\widehat{D}_0} (\sign H_t) - \corr_{\mathcal{D}} (\sign H_t) |\leq\varepsilon/10$, and $\max_{t\in [T]} |\corr_{\widehat{D}'_t} (W_t) - \corr_{\mathcal{D}_t} (W_t) |\leq\gamma\varepsilon/20\mathcal{C}_\mathcal{X}$ can be ensured with probability $1-\delta$ by a simple application of Hoeffing's inequality and union bound given the setting of $m=m(\varepsilon_0,\delta_0)+\mathcal{O}((\mathcal{C}_\mathcal{X})^2/\varepsilon^2\gamma^2)$ and $S_0=\mathcal{O}(1/\gamma^2\varepsilon^2)$. Similarly, $\varepsilon_{\textrm{Gen}}\leq\gamma\varepsilon/10$ holds in \Cref{lem:concmain} for $S = \Omega((\VCB+\log\delta^{-1})/\gamma^2\varepsilon^2)$ and $U = \Omega((\mathcal{C}_\mathcal{X})^2(\VCB+\log\delta^{-1})/\gamma^2\varepsilon^2)$.

Let $h_t = (H_{t+1}-H_t)/\eta$. Since $\psi$ is $1$-smooth, we have
\begin{align}\label{eq:smoothmain3}
	\Phi_{\mathcal{D}}(H_{t+1}) - \Phi_\mathcal{D}(H_t) 
	\leq \eta \Phi'_\mathcal{D}(H_t, h_t) + \frac{\eta^2\mathcal{C}_\mathcal{X}}{2\gamma^2}.
\end{align}

{\bf Case A}: Consider any step $t$ where $\corr_{\widehat{D}'_t}(W_t) >\tau$. Here $h_t = W_t/\gamma$. It follows from \Cref{lem:concmain4} that
\begin{align*}
	\Phi'(H_t, h_t) &\leq -\frac{(1+\mathcal{C}_\mathcal{X})\corr_{\mathcal{D}_t}(h_t)}{\gamma} + \frac{\varepsilon_{\textrm{Gen}}}{\gamma}\\
	&\leq -\frac{(1+\mathcal{C}_\mathcal{X})\corr_{\widehat{D}'_t}(h_t)}{\gamma} + \frac{2\varepsilon_{\textrm{Gen}}}{\gamma} +\frac{\varepsilon}{5}\\
    &\leq -\varepsilon
\end{align*}
where $\tau=2\gamma\varepsilon/(1+\mathcal{C}_\mathcal{X})$ and $\eta =\gamma^2\varepsilon/\mathcal{C}_\mathcal{X}$. By \Cref{eq:smoothmain3}, the potential drops as $\Phi(H_{t+1}) -\Phi(H_t)\leq -\gamma^2\varepsilon^2/2\mathcal{C}_\mathcal{X}$.

{\bf Case B}: Consider any step $t$ where $\corr_{\widehat{D}'_t}(W_t) \leq \tau$ and {\em crucially} $\Phi(H_t, \sign H_t) \geq \varepsilon$. Here $h_t = -\sign H_t$. Since $\Phi'(H_t, h_t) = -\Phi'(H_t, \sign H_t)\leq -\varepsilon$, by \Cref{eq:smoothmain3}, we have $\Phi(H_{t+1}) -\Phi(H_t)\leq -\gamma^2\varepsilon^2/2\mathcal{C}_\mathcal{X}$.

At initialization, $\Phi(\mathbf{0})=0$. Further for any $H:\mathcal{X}\to \mathbb{R}$, using non-negativity of $\psi$, we have
\[ \Phi(H) = \mathcal{C}_\mathcal{X}\E_{x\sim \mathcal{Q}} [\psi(H(x))] - \E_{(x,y)\sim \mathcal{D}}[yH(x)] \geq \E_{x\sim {\mathcal{D}_\mathcal{X}}} [\psi(H(x))] - \E_{(x,y)\sim \mathcal{D}}[yH(x)] \geq \frac{1}{2}.\]
Thus, at initialization $\Phi$ is at most half away from its minimum. Thus, setting $T=2\mathcal{C}_\mathcal{X}/\gamma^2\varepsilon^2$, there must arise an iterate such that neither Case A nor Case B hold. That is, there is some $s\in [T]$ such that $\corr_{\widehat{D}'_s}(W_s) \leq \tau$ and $\Phi_\mathcal{D}(H_s, \sign H_s) \leq \varepsilon$. Now using \Cref{lem:concmain4} and that the weak learner $\gamma$-approximately maximizes correlation (\Cref{def:wl}), we have
\begin{align*}
	-\Phi'(H_s, h^*) &\leq (1+\mathcal{C}_\mathcal{X})\corr_{\mathcal{D}_s}(h^*) + 2\varepsilon_{\textrm{Gen}} \\
	&\leq \frac{ (1+\mathcal{C}_\mathcal{X}) \corr_{\mathcal{D}_s}(W_s) }{\gamma} + \frac{ (1+\mathcal{C}_\mathcal{X}) \varepsilon_0 }{\gamma} + 2\varepsilon_{\textrm{Gen}}\\
	&\leq \frac{ (1+\mathcal{C}_\mathcal{X})\corr_{\widehat{D}'_s}(W_s)}{\gamma} + \frac{(1+\mathcal{C}_\mathcal{X})\varepsilon_0}{\gamma} + \frac{\varepsilon}{5} \\
    &\leq \frac{(1+\mathcal{C}_\mathcal{X})\varepsilon_0}{\gamma} + \frac{12\varepsilon}{5}
\end{align*}
where in the last line we recall $\tau=2\varepsilon\gamma/(1+\mathcal{C}_\mathcal{X})$ and $\varepsilon_{\textrm{Gen}}=\varepsilon\gamma/10$.

By \Cref{lem:consmain}, we have 
\begin{align*}
	\corr_\mathcal{D}(h^*) - \corr_\mathcal{D}(\sign(H_s)) &\leq \Phi'_\mathcal{D}(H_s, \sign(H_s)) - \Phi'_\mathcal{D}(H_s, h^*) \\
    &\leq \frac{(1+\mathcal{C}_\mathcal{X})\varepsilon_0}{\gamma} + \frac{17\varepsilon}{5}.
\end{align*}

To complete the proof, we observe that
\begin{align*}
	\corr_\mathcal{D}(\overline{h}) &\leq \corr_{\widehat{D}_0}(\overline{h}) + \varepsilon/10\\
	&\leq \corr_{\widehat{D}_0}(\sign H_s) + \varepsilon/10\\
	&\leq \corr_{\mathcal{D}}(\sign H_s) + \varepsilon/5 \\
    &\leq (1+\mathcal{C}_\mathcal{X})\varepsilon_0/\gamma + 18\varepsilon/5,
\end{align*}
where we use the fact that $\overline{h}$ maximizes the empirical correlation on the dataset $\widehat{D}_0$.
\end{proof}

%% file: icml/experiment_app.tex
\section{Additional experimental details}\label{sec:experiment_app}

For PAB, the number of samples that can be fed to a week learner in a round scales inversely with the number of boosting rounds, as the algorithm requires fresh samples each round.As such, we perform a grid search on the number of boosting rounds with $T \in \{25, 50, 100\}$, while we just use $100$ for our implementation of \cref{alg:rev}. In both algorithms we search over the parameter $m$, the number of samples we feed to the weak learner each round with a grid of $\{5,20, 50, 100\}$, though if such a setting is invalid for PAB, we continue until all samples are used.

 Our experiments were performed using the fractional relabeling scheme stated in \cite{kanade2009potential}, intended to reduce the stochasticity the algorithm is subject to. In particular, rather than sampling labels, we provide both $(x, y)$ and $(x, -y)$ in our dataset with weights equal to their sampling probabilities. Experiments are all run on an M1 Macbook Pro and complete within an hour. Multiclass datasets are converted to binary.

%% file: icml/app_applications.tex
\section{Proof of \cref{thm:half}}\label{sec:proof_half}
\begin{proof}
We observe that ERM on the Fourier basis $\chi_S(x) = \prod_{i \in S} x_i$, namely parities on subsets $S$, can be used to produce a weak learner \cite{klivans2004learning}. 
As such, an $n$-dimensional halfspace can be approximated with uniform weighting on the hypercube
to $\varepsilon^2$ $\ell_2$-error using degree-limited $\mathcal{B}_{n,d} = \{\pm \chi_S: |S| \le d \}$ as a basis, where $d = 20 \varepsilon^{-4}$.
As a result, at least one $h \in \mathcal{B}_{n,d}$ must have high correlation. 

\begin{lemma}[Lemma 5 in \cite{kalai2008agnostic}]\label{lem:half_weak}
    Let $\mathcal{D}$ be any data distribution over $ \{\pm 1\}^n \times \{-1, 1\}$ with marginal distribution $\text{Unif}(\{\pm 1\}^n)$ on the features. For any fixed $\varepsilon$ and $d = 20 \varepsilon^{-4}$, there exists some $h \in \mathcal{B}_{n,d}$ such
    \begin{align*}
        \corr_{\mathcal{D}}(h) \geq \frac{\max_{c \in \mathcal{H}} \corr_{\mathcal{D}}(c) - \varepsilon}{n^d}
    \end{align*}
\end{lemma}

The result follows directly from the preceding lemma, which provides a weak learner for the task, and \Cref{thm:main}. We note that $|\mathcal{B}_{n,d}| < n^d$ and $\gamma = n^{-d}$, so 
    $$ \frac{\log |\mathcal{B}_{n,d}|}{\gamma^2\varepsilon^2} \leq \frac{dn^{2d}\log(n)}{\varepsilon^2}. $$
The unlabeled samples used in \cref{alg:rev} can be produced by sampling from the hypercube adding to the $n^{\mathrm{poly}(1/\varepsilon)}$ runtime, but not the sample complexity.
\end{proof}

\section{Boosting for reinforcement learning}\label{app:rl}
In this section, we consider boosting in the reinforcement learning setting. We wish to separately consider the number of reward-annotated episodes against the number of reward-free episodes needed to learn a near-optimal policy.

Consider a Markov Decision Process $\mathcal{M}=(\mathcal{S}, \mathcal{A}, r, P, \beta, \mu_0)$, where $\mathcal{S}$ is a set of states, $\mathcal{A}=\{\pm 1\}$ is a binary set of actions, $r:\mathcal{S}\times \mathcal{A}\to [0,1]$ determines the (expected) reward at any state-action pair (which is sometimes available), $P:\mathcal{S}\times \mathcal{A}\to \mathcal{S}$ captures the transition dynamics of the MDP, i.e., $P(s'|s,a)$ is the probability of moving to state $s'$ upon taking action $a$ at state $s$, $\beta\in [0,1)$ is the discount factor, and $\mu_0$ is the initial state distribution. Let $Q^\pi(s,a)$ and $V^\pi(s)$ be the state-action and state value functions. Let $V^\pi_\mu=\mathbb{E}_{s\sim \mu}[V(s)]$ be the expected total reward when starting from the start state distribution $\mu$, and we will say $V^\pi_{\mu_0}=V^\pi$. Finally, the occupancy measure $\mu^\pi_{\mu'}$ induced by a policy $\pi$ starting from an initial state distribution $\mu'$ is stated below. We will take $\mu^\pi = \mu^\pi_{\mu_0}$ as a matter of convention.

In the {\em episodic model}, the learner interacts with the MDP in a limited number of episodes of reasonable length (i.e., $\approx (1-\beta)^{-1}$), and the starting state of MDP is always drawn from $\mu_0$. In the second, termed {\em rollouts with $\nu$-resets}, the learner's interaction is still limited to a small number of episodes, however, the MDP now samples its starting state from $\nu$. It is important to stress that in both cases, the learner's objective is the same, to maximize $V^\pi$ starting from $\mu_0$. However, $\nu$ could be more {\em spread out} over the state space than $\mu_0$, and provide an implicit source of explanation, and the learner's guarantee as shown next benefits from its dependence on a milder notion of distribution mismatch in this case. In this setting, we do not always assume the reward is revealed. We consider a model where we can rollout a policy and observe rewards or alternatively can just observe the state trajectories.

Since we have binary actions, our weak learners are policies, which we denote $\pi$ instead of $h$. This notion is equivalent to that used by \citet{brukhim2022boosting} and \citet{ghai2024sampleefficient}, because for binary actions, a random policy induces an accuracy of half regardless of the distribution over features and labels.

Say $\pi^*\in\argmax_{\pi} V^\pi$ be a reward maximizing policy, and $V^*$ be its value. Let $\mathbbl{\Pi}$ be the convex hull of the boosted policy class, i.e., the outputs of the boosting algorithm. For any state distribution $\mu'$, define the policy completeness $\mathcal{E}_{\mu'}$ term as 
$$ \mathcal{E}_{\mu'} = \max_{\pi\in \mathbbl{\Pi}} \min_{\pi'\in \Pi} \mathbb{E}_{s\sim \mu^\pi_{\mu'}} [\max_{a\in \mathcal{A}} Q^\pi(s,a) - \mathbb{E}_{a\sim \pi'(\cdot|s)}Q^\pi(s,a)].$$
In words, this term captures how well the greedy policy improvement operator is approximated by $\Pi$ in an state-averaged sense over the  distribution induces by any policy in $\mathbbl{\Pi}$. Finally, we define distribution mismatch coefficients below.
$$ C_\infty = \max_{\pi\in \mathbbl{\Pi}} \|\mu^{\pi^*}/\mu^\pi\|_\infty, \quad D_\infty = \|\mu^{\pi^*}/\nu\|_\infty. $$

\begin{algorithm}[t]
    \caption{RL Boosting adapted from \cite{brukhim2022boosting}}\label{alg:rlMAIN1}
    \begin{algorithmic}[1]
        \STATE \textbf{Input}: iteration budget $T$, state distribution $\mu$, step sizes $\eta_{t}$, post-selection sample size $P$
        \STATE Initialize a policy $\pi_0\in \Pi$ arbitrarily. 
        \FOR{$t=1$ {\bfseries to} $T$}
        \STATE Run \Cref{alg:rev2} to get $\pi'_t$, using 
        \begin{itemize}
            \item \Cref{alg:q_sampler} to produce a distribution over state-actions (ignore $\widehat{Q}$) by executing the current policy $\pi_{t-1}$ starting from the initial state distribution $\mu$ as the labeled samples.
            \item \Cref{alg:q_sampler_rfree} to produce a distribution over states by executing the current policy $\pi_{t-1}$ starting from the initial state distribution $\mu$ as the unlabeled samples.
        \end{itemize}
        \STATE Update $\pi_t = (1-\eta_{t})\pi_{t-1} + \eta_{t} \pi'_t$.
        \ENDFOR
    \STATE Run each policy $\pi_t$ for $P$ rollouts to compute an empirical estimate $\widehat{V^{\pi_t}}$  of the expected return.
    \STATE \textbf{return} $\overline{\pi} = \pi_{t'}$ where $t'=\argmax_t \widehat{V^{\pi_t}}$.
    \end{algorithmic}
\end{algorithm}

\begin{algorithm}[t]
    \begin{algorithmic}[1]
        \STATE Sample state $s_0 \sim \mu$ and action  $a' \sim \text{Unif}(\mathcal{A})$.
        \STATE Sample $s\sim \mu^{\pi}$ as follows:
        at every step $h$,  with probability $\beta$, execute $\pi$; else, accept  $s_h$. 
        \STATE\label{state:next_step} Take action $a'$ at state $s_h$, then
        continue to execute $\pi$, and use a termination probability of
        $1-\beta$. Upon termination, set
        $R(s_h,a')$ as the sum
        of rewards from time $h$ onwards.
        \STATE Define the vector $\widehat{Q}$, such that for all $a \in A$, $\widehat{Q}(a) = 2R(s_h,a') \cdot \mathbb{I}_{a=a'}$.
        \STATE With probability $C\widehat{Q}(a')$, set $y=a'$ else set $y\in \mathcal{A}-\{a'\}$, where $C=(1-\beta)/2$.
        \STATE \textbf{return} $(s_h, \widehat{Q}, y)$.
    \end{algorithmic}
    \caption{Trajectory Sampler adapted from \cite{brukhim2022boosting}}
    \label{alg:q_sampler}
\end{algorithm}

\begin{algorithm}[t]
    \begin{algorithmic}[1]
        \STATE Sample state $s_0 \sim \mu$ and action  $a' \sim \text{Unif}(\mathcal{A})$.
        \STATE Sample $s\sim \mu^{\pi}$ as follows:
        at every step $h$,  with probability $\beta$, execute $\pi$; else, accept  $s_h$. 
        \STATE\label{state:next_step2} Take action $a'$ at state $s_h$, then
        continue to execute $\pi$, and use a termination probability of
        $1-\beta$. 
        \STATE \textbf{return} $s_h$.
    \end{algorithmic}
    \caption{Reward-free Trajectory Sampler}
    \label{alg:q_sampler_rfree}
\end{algorithm}

\begin{theorem}\label{thm:rl2}
Let $\mathcal{W}$ be a $\gamma$-weak learner for the policy class $\Pi$ operating with a base class $\mathcal{B}$, with sample complexity $m(\varepsilon_0, \delta_0) = (\log |\mathcal{B}|/\delta_0)/\varepsilon_0^2$. Fix tolerance $\varepsilon$ and failure probability $\delta$. In the {\em episodic} access model, there is a setting of parameters such that \Cref{alg:rlMAIN1} when given access to $\mathcal{W}$ produces a policy $\overline{\pi}$ such that with probability $1-\delta$, we have
$$ V^* - V^{\overline{\pi}} \leq \frac{C_\infty \mathcal{E}}{1-\beta} + \varepsilon,$$
while sampling $$\mathcal{O}\left(\frac{C_\infty^5\log |\mathcal{B}|}{(1-\beta)^9 \gamma^3 \varepsilon^4}\right)$$ episodes of length $\mathcal{O}((1-\beta)^{-1})$ without reward feedback (via \cref{alg:q_sampler_rfree}) and 
$$\mathcal{O}\left(\frac{C_\infty^4\log |\mathcal{B}|}{(1-\beta)^7 \gamma^2 \varepsilon^3}\right)$$ episodes of length $\mathcal{O}((1-\beta)^{-1})$ with reward feedback (via \cref{alg:q_sampler}).

In the {\em $\nu$-reset} access model, there is a setting of parameters such that \Cref{alg:rlMAIN1} when given access to $\mathcal{W}$ produces a policy $\overline{\pi}$ such that with probability $1-\delta$, we have
$$ V^* - V^{\overline{\pi}} \leq \frac{D_\infty \mathcal{E}_\nu}{(1-\beta)^2} + \varepsilon,$$
while sampling $$\mathcal{O}\left(\frac{D_\infty^5\log |\mathcal{B}|}{(1-\beta)^{15}\gamma^3 \varepsilon^5}\right)$$ episodes of length $\mathcal{O}((1-\beta)^{-1})$
 without reward feedback (via \cref{alg:q_sampler_rfree}) and 
$$\mathcal{O}\left(\frac{D_\infty^4\log |\mathcal{B}|}{(1-\beta)^{12}\gamma^2 \varepsilon^4}\right)$$ episodes of length $\mathcal{O}((1-\beta)^{-1})$ with reward feedback (via \cref{alg:q_sampler}).
\end{theorem}

\begin{proof}
The proof follows by applying the result in \cref{thm:main2} within the proof of Theorem~22 from \citet{ghai2024sampleefficient}.

For the episodic model, applying the second part of Theorem 9 in \cite{brukhim2022boosting}, while noting the smoothness of $V^\pi$, and combining the result with Lemma 18 and Lemma 11 in \cite{brukhim2022boosting}, we have with probability $1-T\delta$

Following the logic of Theorem~22 from \citet{ghai2024sampleefficient}, we need to ensure is that output of \Cref{alg:rev2} as instantiated in \Cref{alg:rlMAIN1} every round has an excess correlation gap over the best policy $\Pi$ no more that $(1-\beta)^2\varepsilon/C_\infty$, which \Cref{alg:rev2} assures us can be accomplished with $\mathcal{O}\left(\frac{C_\infty^3 \log |\mathcal{B}|}{(1-\beta)^6\gamma^3 \varepsilon^3}\right)$ unlabeled samples and  $\mathcal{O}\left(\frac{C_\infty^2 \log |\mathcal{B}|}{(1-\beta)^4\gamma^2 \varepsilon^2}\right)$ labeled samples. The total number of samples is $T=\mathcal{O}\left(\frac{C_\infty^2}{(1-\beta)^3 \varepsilon}\right)$ times greater.

Similarly, for the $\nu$-reset model, we need to ensure is that output of \Cref{alg:rev2} as instantiated in \Cref{alg:rlMAIN1} every round has an excess correlation gap over the best policy $\Pi$ no more that $(1-\beta)^3\varepsilon/D_\infty$, which \Cref{alg:rev2} assures us can be accomplished with $\mathcal{O}\left(\frac{D_\infty^3 \log |\mathcal{B}|}{(1-\beta)^9\gamma^3 \varepsilon^3}\right)$ unlabeled samples and $\mathcal{O}\left(\frac{D_\infty^2 \log |\mathcal{B}|}{(1-\beta)^6\gamma^2 \varepsilon^2}\right)$ labeled samples. The total number of samples is $T=\mathcal{O}\left(\frac{D_\infty^2}{(1-\beta)^6 \varepsilon^2}\right)$ times greater.
\end{proof}